\documentclass[10pt,twocolumn,letterpaper]{article}
\usepackage{cvpr} 
\usepackage{booktabs}
\usepackage{multirow}
\usepackage{bm}
\usepackage{amsmath}
\usepackage[table]{xcolor}

\newcount\K
\def\bla#1{
\K=0 \loop\ifnum\K<#1
{\textcolor[gray]{0.9}{{\it bla bla bla bla bla bla bla bla bla bla bla bla bla bla bla}}}
\advance\K by1\repeat
}

\usepackage[utf8]{inputenc}
\definecolor{cvprblue}{rgb}{0.21,0.49,0.74}
\usepackage[pagebackref,breaklinks,colorlinks,allcolors=cvprblue]{hyperref}
\def\paperID{1357}
\def\confName{CVPR}
\def\confYear{2026}

\def\paragraph#1{\noindent \textbf{#1}}

\title{BioVITA:
Biological Dataset, Model, and Benchmark\\
for Visual-Textual-Acoustic Alignment}
\vspace{-10pt}
\author{
Risa Shinoda$^{1,2}$ \quad
Kaede Shiohara$^2$ \quad
Nakamasa Inoue$^3$ \quad
\\[0.8em]
Kuniaki Saito$^4$ \quad
Hiroaki Santo$^1$ \quad
Fumio Okura$^1$
\\[0.8em]
\normalsize{
$^1$The University of Osaka \quad
$^2$The University of Tokyo \quad
$^3$Institute of Science Tokyo \quad
$^4$OMRON SINIC X
}}
\begin{document}
\maketitle
\begin{abstract}
Understanding animal species from multimodal data poses an emerging challenge at the intersection of computer vision and ecology.
While recent biological models, such~as BioCLIP, have demonstrated strong alignment between images and textual taxonomic information for species identification, the integration of the audio modality remains an open problem.
We propose \scalebox{0.95}{$\mathsf{BioVITA}$}, a novel visual-textual-acoustic alignment framework for biological applications.
\scalebox{0.95}{$\mathsf{BioVITA}$} involves (i) a training dataset, (ii) a representation model, and (iii) a retrieval benchmark.
First, we construct a large-scale training dataset comprising 1.3 million audio clips and 2.3 million images, covering 14,133 species annotated with 34 ecological trait labels.
Second, building upon BioCLIP~2, we introduce a two-stage training framework to effectively align audio representations with visual and textual representations.
Third, we develop a cross-modal retrieval benchmark that covers all possible directional retrieval across the three modalities (i.e., image-to-audio, audio-to-text, text-to-image, and their reverse directions), with three taxonomic levels: Family, Genus, and Species.
Extensive experiments demonstrate that our model learns a unified representation space that captures species-level semantics beyond taxonomy, advancing multimodal biodiversity understanding. The project page is available at: \url{https://dahlian00.github.io/BioVITA_Page/}
\end{abstract}

\section{Introduction}
\label{sec:intro}

\begin{figure}[t]
\centering
\includegraphics[width=1.0\linewidth]{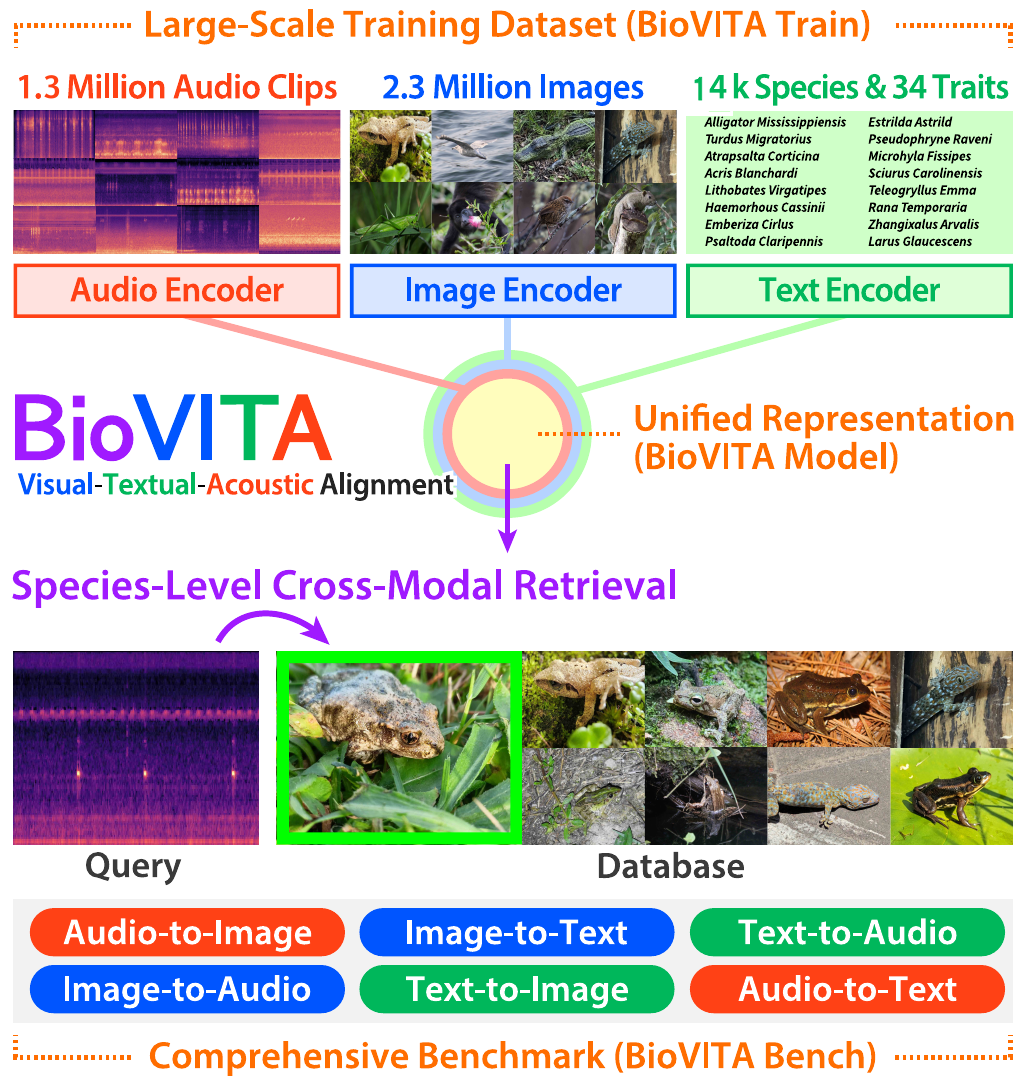}
\vspace{-16pt}
\caption{
We introduce \scalebox{0.95}{$\mathsf{BioVITA}$} for biological visual-textual-acoustic alignment.
We curate a million-scale dataset,
train a unified representation model, and develop a comprehensive species-level retrieval benchmark.
}
\vspace{-18pt}
\label{fig:teaser_animalclap}
\end{figure}

Biological vision models have become essential for understanding animal behavior and ecosystem dynamics, integrating insights from computer vision and ecology.
Inspired by visual-textual alignment frameworks such as CLIP~\cite{clip},
BioCLIP~\cite{stevens2024bioclip, Gu2025BioCLIP2} has recently established alignment of biological images with a hierarchical taxonomy represented by structured text prompts,
achieving impressive zero-shot species identification performance.
Similarly, in the audio domain, CLAP~\cite{laionclap2023} has introduced acoustic pre-training with text data analogous to CLIP, leading to several follow-up studies focusing on animal vocalizations~\cite{robinson2025naturelm,bioacoustics}.
Despite these advances, \textit{visual-textual-acoustic (VITA)} alignment for integrating image, taxonomic text, and audio representations remains an open challenge.
As biodiversity research often relies on perceiving species through complementary sensory modalities, 
achieving effective integration is crucial for more comprehensive species understanding.

To establish VITA alignment, a dataset perspective is indispensable for both training and evaluation.
However, current multimodal datasets primarily focus only on pairwise modalities, either image-text~pairs\cite{vendrow2024inquire,Horn2017TheIS,Gu2025BioCLIP2,seaturtle,Cermak_2024_WACV,fishnet,birdnet}, or audio-text pairs ~\cite{Hagiwara2022BEANSTB,bioacoustics}.
Because these datasets often differ in their taxonomic hierarchies and overall scale, there is a need for a comprehensive multimodal training and evaluation dataset that unifies all modalities within a consistent ecological context.

Motivated by these limitations, we introduce \scalebox{0.95}{$\mathsf{BioVITA}$}, a novel VITA alignment framework comprising
(i) a million-scale training dataset (\scalebox{0.95}{$\mathsf{BioVITA\hspace{1.5pt}Train}$}), (ii) a unified representation model (\scalebox{0.95}{$\mathsf{BioVITA\hspace{1.5pt}Model}$}),
and (iii) a species-level cross-modal retrieval benchmark (\scalebox{0.95}{$\mathsf{BioVITA\hspace{1.5pt}Bench}$}).
As shown in Fig.~\ref{fig:teaser_animalclap}, our model consists of audio, image, and text encoders trained on the dataset involving 1.3 million audio clips and 2.3 million images spanning 14k species.
After learning unified representations, the model is evaluated across six comprehensive retrieval directions: image-to-audio, audio-to-text, text-to-image, and their reverse directions.
This framework advances multimodal biodiversity understanding.
In summary, our contributions are threefold.

\begin{enumerate}
\item We introduce \scalebox{0.95}{$\mathsf{BioVITA\hspace{1.5pt}Train}$} (\S\ref{sec:training_dataset}), a training dataset for VITA alignment. We curate 1.3 million audio clips and 2.3 million images
with textual taxonomic annotations, covering 14k species and 34 ecological traits.
\item We propose \scalebox{0.95}{$\mathsf{BioVITA\hspace{1.5pt}Model}$} (\S\ref{sec:model}), a unified representation model.
Through two-stage training, our model effectively aligns audio representations with pre-trained visual and textual representations.
\item We develop \scalebox{0.95}{$\mathsf{BioVITA\hspace{1.5pt}Bench}$} (\S\ref{sec:benchmark}), a species-level retrieval benchmark spanning the six cross-modal directions. Our benchmark enables comprehensive analysis from multimodal, ecological, and generalization perspectives.
\end{enumerate}

\section{Related Works}
\label{sec:related}

\paragraph{Species Recognition from Images.}
Animals exhibit distinctive visual characteristics across species, making fine-grained visual recognition an important research topic. 
Numerous datasets and models have contributed to this field
~\cite{van2018inaturalist,wah2011caltech,10648043,yu2021ap,ng2022animal,shinoda2025petface,shinoda2025animalclue}
, as well as fine-grained classification models such as B-CNN~\cite{lin2015bilinear}, multi-attention~\cite{sun2018multi}, Cross-X~\cite{luo2019cross}, and TransFG~\cite{he2022transfg}.
Recently, BioCLIP~\cite{stevens2024bioclip} and BioCLIP2~\cite{Gu2025BioCLIP2} have explored image-text representation learning, 
which has significantly advanced cross-domain understanding in biodiversity.

\paragraph{Species Recognition from Audio.}
Recent advances in acoustic sensing technologies, particularly the deployment of automated recording units (ARUs), have enabled large-scale, continuous monitoring of natural environments, underscoring the growing importance of bioacoustic analysis in ecological research~\cite{acoustic_monitoring_review,ecological_audio_review,stowell2022computational}.
Building on these developments, recent work in signal processing and machine learning has made substantial progress in automated species recognition from acoustic signals~\cite{Transferable,robinson2025naturelmaudio,inatsound,beans,Wood2022BirdNET,shinoda2026animalclap}.
For example, BioLingual~\cite{Transferable} demonstrated the effectiveness of linking animal vocalizations with textual representations via contrastive language–audio pretraining (CLAP)~\cite{laionclap2023}, achieving state-of-the-art results in species classification and detection.
Similarly, NatureLM-Audio~\cite{robinson2025naturelmaudio} extended large-scale multimodal learning to acoustic ecology, supporting cross-species retrieval and sound-based biodiversity indexing.
Other frameworks such as BirdNET~\cite{Wood2022BirdNET} and Perch~\cite{van2025perch} have further advanced robust detection and identification pipelines for large-scale bird monitoring, collectively illustrating how foundation audio models and ecoacoustic datasets are transforming species-level recognition and ecological monitoring.

\definecolor{lightcyan}{rgb}{0.88,0.95,1}
\begin{table}[t]
\centering 
\small
\caption{Comparison with existing animal sound datasets.
The number of species excludes subspecies.
Com and Sci stand for common and scientific names, respectively.
}
\label{tab:datasets}
\vspace{-8pt}
\resizebox{\linewidth}{!}{
\setlength{\tabcolsep}{1pt}
\begin{tabular}{lccccc} 
\toprule
    {Dataset}  & {\#Audio Clips} & {\#Images} & {Taxon} & {\#Species} & {\#Traits}  \\
\midrule
BEANS~\cite{Hagiwara2022BEANSTB} & 56k & 0 & Com & 391 & 0 \\
AnimalSpeak~\cite{bioacoustics} & 1.1M & 0 & Com, Sci & 12k & 0 \\
iNatSounds~\cite{inatsounds} & 230k & 0 & Full & 5.5k & 0 \\
SSW60~\cite{ssw60} & 4k & 31k & Com & 60 & 0 \\
\rowcolor{lightcyan} 
\scalebox{0.95}{$\mathsf{BioVITA\hspace{1.5pt}Train}$} & 1.3M & 2.3M & Full & 14k & 34\\[-0.2em]
\bottomrule
\end{tabular}
}
\vspace{-15pt}
\end{table}

\paragraph{Multi-modal Recognition.}
In contrast to the two domains above, research bridging visual and acoustic modalities remains limited.
SSW60~\cite{ssw60} is a pioneering study integrating video, audio, and image modalities for bird classification, but it is limited to 60 species.
Recent advances in large-scale multi-modal representation learning have demonstrated that unified embeddings across modalities, such as image, audio, and text, can greatly enhance cross-domain generalization~\cite{dhakal2024sat2cap,10642394}.
ImageBind~\cite{girdhar2023imagebind}, for instance, learns a shared embedding space across six modalities without pairwise supervision, enabling strong zero-shot transfer.
Building on this paradigm, several works have begun applying multimodal foundation models to ecology and biodiversity monitoring~\cite{robinson2025naturelmaudio,Transferable,laionclap2023,kryklyvets-etal-2025-mavis}, linking animal vocalizations, textual descriptions, and visual cues into a unified semantic space\footnote{Some of these models are trained on datasets that overlap with our benchmarks, raising the possibility of test-time data leakage. We do not directly benchmark against these models.}\looseness=-1.
TaxaBind~\cite{sastry2025taxabind} also contributes to extending this multi-domain training paradigm into the animal domain. While it adopts a similar joint embedding approach, it only bridges the image modality. In addition, TaxaBind is trained on a relatively small audio dataset of 75k samples.
Inspired by these efforts, we extend multimodal learning to a broader ecological setting, jointly modeling animal appearance and sound to support cross-species generalization and behavioral understanding in the wild.

\paragraph{Dataset Comparison.}
Table~\ref{tab:datasets} summarizes dataset statistics with a comparison to existing animal vocalization datasets.
As shown, \scalebox{0.95}{$\mathsf{BioVITA\hspace{1.5pt}Train}$} is the largest tri-modal dataset in terms of scale, comprising over one million samples for both audio and visual modalities, further enriched by detailed ecological trait annotations.

\section{Training Dataset for BioVITA}
\label{sec:training_dataset}

We introduce \scalebox{0.95}{$\mathsf{BioVITA\hspace{1.5pt}Train}$},\hspace{2pt}a large-scale training dataset for VITA alignment within a unified ecological taxonomy.
The dataset consists of 1.3 million audio clips and 2.3 million images with their textual labels covering 14k species (excluding subspecies) and 34 fine-grained traits.
All data are collected from publicly available sources under a consistent and license-compatible protocol.

\subsection{Dataset Construction}
\label{sec:dataset_construction}

{\renewcommand{\arraystretch}{0.85}
\begin{table}[t]
\centering
\caption{Ecological trait labels for \scalebox{0.95}{$\mathsf{BioVITA}$}. $^{\star}$ indicates mutually exclusive traits, where only one trait within a category is active.}
\label{tab:trait_annotation}
\vspace{-4pt}
\begin{minipage}[t]{0.449\linewidth}
\centering
\footnotesize
\setlength{\tabcolsep}{8.5pt}
\begin{tabular}{ll}
\toprule
\textbf{Category} & \textbf{Trait} \\[-0.1em]
\midrule
\multirow{4}{*}{\rotatebox{0}{\shortstack{Diet Type$^{\star}$}}} & Herbivorous \\
 & Carnivorous \\
 & Omnivorous \\
 & Specialized \\[-0.17em]
\midrule
\multirow{4}{*}{\rotatebox{0}{\shortstack[l]{Activity\\Pattern$^{\star}$}}} & Diurnal \\
 & Nocturnal \\
 & Crepuscular \\
 & Cathemeral \\[-0.16em]
\midrule
\multirow{3}{*}{\rotatebox{0}{\shortstack[l]{Locomotion\\Posture$^{\star}$}}} & Quadrupedal \\
 & Bipedal \\
 & Other \\[-0.21em]
\midrule
\multirow{5}{*}{\rotatebox{0}{\shortstack{Lifestyle}}} & Arboreal \\
 & Aquatic \\
 & Terrestrial \\
 & Fossorial \\
 & Aerial \\[-0.11em]
\midrule
\multirow{1}{*}{\rotatebox{0}{Trophic Role}} & Predator \\
\bottomrule
\end{tabular}
\end{minipage}
\hspace{14pt}
\begin{minipage}[t]{0.449\linewidth}
\footnotesize
\setlength{\tabcolsep}{8.5pt}
\begin{tabular}{ll}
\toprule
\textbf{Category} & \textbf{Trait} \\[-0.1em]
\midrule
\multirow{6}{*}{\rotatebox{0}{Habitat}} & Forest \\
 & Grassland \\
 & Desert \\
 & Wetland \\
 & Mountain \\
 & Urban \\
\midrule
\multirow{5}{*}{\rotatebox{0}{\shortstack[l]{Climatic\\Distribution}}} & Tropical \\
 & Subtropical \\
 & Temperate \\
 & Boreal \\
 & Polar \\
\midrule
\multirow{4}{*}{\rotatebox{0}{\shortstack[l]{Social\\Behavior$^{\star}$}}} & Solitary \\
 & Pairing \\
 & Grouping \\
 & Herding \\
\midrule
\multirow{2}{*}{\rotatebox{0}{\shortstack[l]{Migration\\Status}}} & Migratory \\
& Resident\\
\bottomrule
\end{tabular}
\end{minipage}
\vspace{-8pt}
\end{table}}

While several prior studies developed training datasets linking images to ecological taxonomies~(\textit{e.g.},~\cite{stevens2024bioclip}), alignment with audio and taxonomic information remains unexplored.
As such, we focus primarily on the audio modality by first curating bioacoustic data.
Specifically, we constructed \scalebox{0.95}{$\mathsf{BioVITA\hspace{1.5pt}Train}$} through three steps: 1) audio data curation, 2) fine-grained annotation, and 3) visual data consolidation.
This pipeline ensures comprehensive coverage and effective multimodal integration with consistent annotations.

\paragraph{1) Audio Data Curation.}\hspace{4pt}To guarantee audio data quality, we curate recordings from three reliable platforms:
iNaturalist~\cite{inaturalist2025}, Xeno-Canto (XC)~\cite{xenocanto2025}, and Animal Sound Archive (ASA)~\cite{animalsoundarchive2025}.
iNaturalist and XC are citizen science platforms that host community-contributed wildlife observations with spatiotemporal metadata.
ASA is a research repository maintained by the Museum f\"ur Naturkunde Berlin, providing archival-quality recordings with expert taxonomic validation.
In total, 1.3 million audio clips are collected under Creative Commons licenses.

\paragraph{2)\hspace{2pt}Fine-Grained\hspace{2pt}Annotation.}\hspace{2pt}
We annotate each audio clip with hierarchical taxonomic labels, including class, order, family, and genus, based on the species information from each platform.
To enable fine-grained analysis, we assign trait labels for 34 ecological traits listed in Table~\ref{tab:trait_annotation}.
These traits cover major ecological categories, such as diet type, activity pattern, and habitat, which are potentially associated with acoustic and visual characteristics~\cite{Morton1975Avian,Anthropogenic,McComb2005Primate,social_complexity,Freeberg2006Chickadees}.
Trait labels were first extracted from iNaturalist webpages using an LLM (GPT-5~\cite{OpenAI2025GPT5}).
We then asked GPT-5 to fill in missing traits and review the completed annotations; any changed values were manually verified.
At this stage, we reserve all data from 325 species that had relatively few samples during the training data collection, and we hold them out from training, together with an additional 10\% of data randomly sampled from all remaining species, to construct \scalebox{0.95}{$\mathsf{BioVITA\hspace{1.5pt}Bench}$} for performance evaluation.

\paragraph{3) Visual Data Consolidation.}
Finally, we integrate visual data into our dataset. Specifically, to align with the species included in our audio dataset, we utilize a corresponding subset of the ToL-200M~\cite{Gu2025BioCLIP2,treeoflife_200m} dataset, an extensive biological image collection aggregating multiple sources.
We randomly sampled 200 images per species, resulting in an image subset comprising 2.3 million images.
Additionally, for benchmarking purposes, we curate a distinct set of 128,645 images from iNaturalist that do not overlap with the ToL-200M dataset.
Please refer to the supplemental material for more information.

\subsection{Statistics and Examples}

\begin{figure}
\centering
\includegraphics[width=\linewidth]{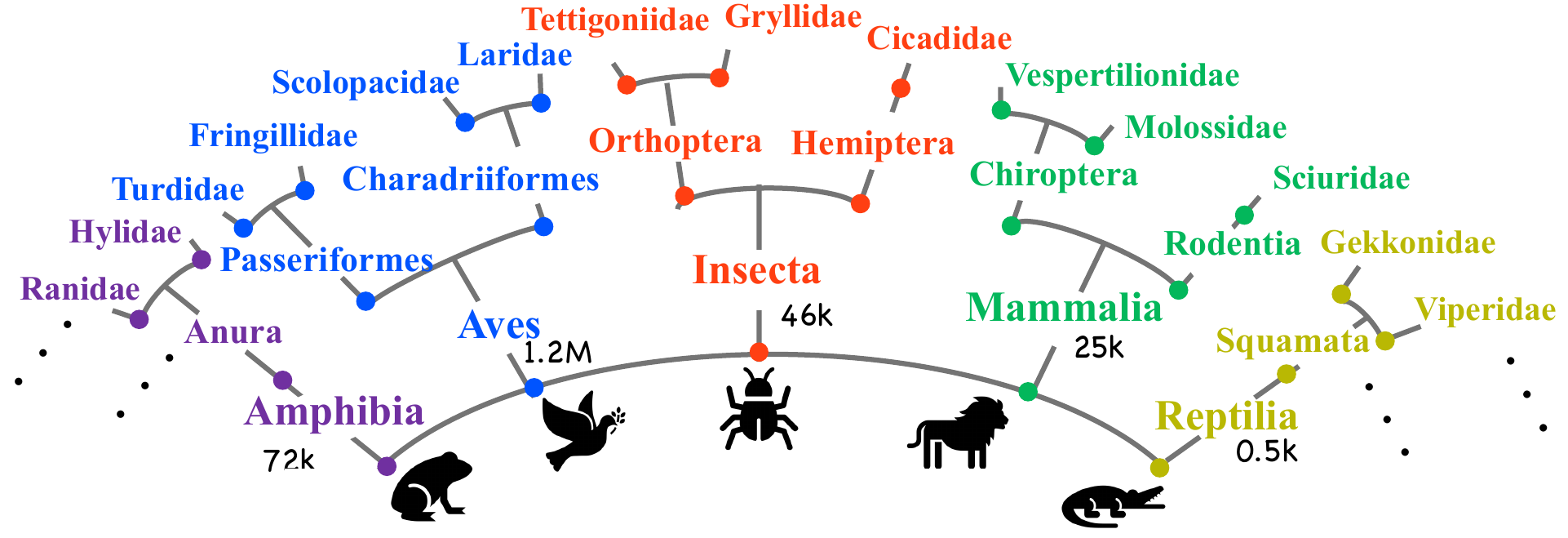}\vspace{-2mm}
\caption{Representative taxonomic distribution of the dataset across the five animal classes. Numbers indicate audio recordings. }

\label{fig:taxonomy}
\end{figure}
\begin{figure}[t]
\centering
\begin{minipage}[b]{0.48\linewidth}
\centering
\includegraphics[width=\linewidth]{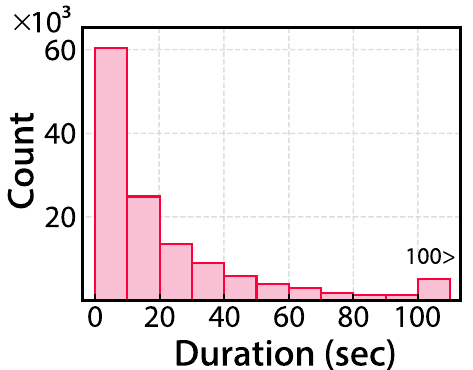}\vspace{-2mm}
\caption{Audio duration.}
\vspace{-5mm}
\label{fig:audio_durations}
\end{minipage}
\hspace{0.01\linewidth}
\begin{minipage}[b]{0.48\linewidth}
    \centering
    \includegraphics[width=\linewidth]{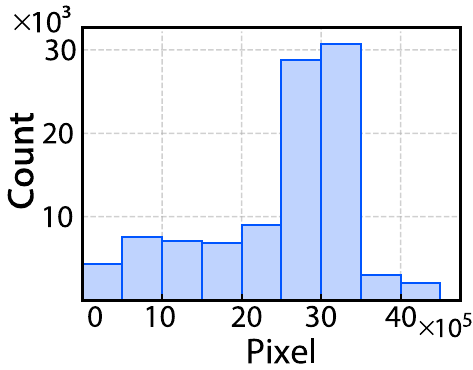}\vspace{-2mm}
    \caption{Image size.}
    \vspace{-5mm}
    \label{fig:image_resolution}
  \end{minipage}
\end{figure}

\paragraph{Taxonomy.}
Our dataset covers 5 distinct classes, 84 orders, 538 families, 3,612 genus, and 14,133 species, underscoring its extensive taxonomic breadth.
As shown in Fig.~\ref{fig:taxonomy}, four acoustically prominent classes are predominant, with \textit{Aves} (birds) exhibiting the greatest diversity, followed by \textit{Amphibia} (amphibians), \textit{Insecta} (insects), and \textit{Mammalia} (mammals).
This comprehensive taxonomic coverage enables detailed ecological modeling.

\paragraph{Audio Duration.}
Figure~\ref{fig:audio_durations} shows the distribution of audio clip durations.
The average duration is 24.6 seconds, indicating sufficient temporal length for capturing characteristic ecological and behavioral signals across species.
Sampling rates are predominantly standardized at 44.1 kHz, ensuring high-fidelity audio suitable for detailed analysis.

\paragraph{Image Size.}
Figure~\ref{fig:image_resolution} presents the distribution of image dimensions. The majority of images exhibit resolutions ranging from 119×119 to 2048×2048 pixels, ensuring ample spatial detail for accurate species identification.

\begin{figure}
\centering
\includegraphics[width=\linewidth]{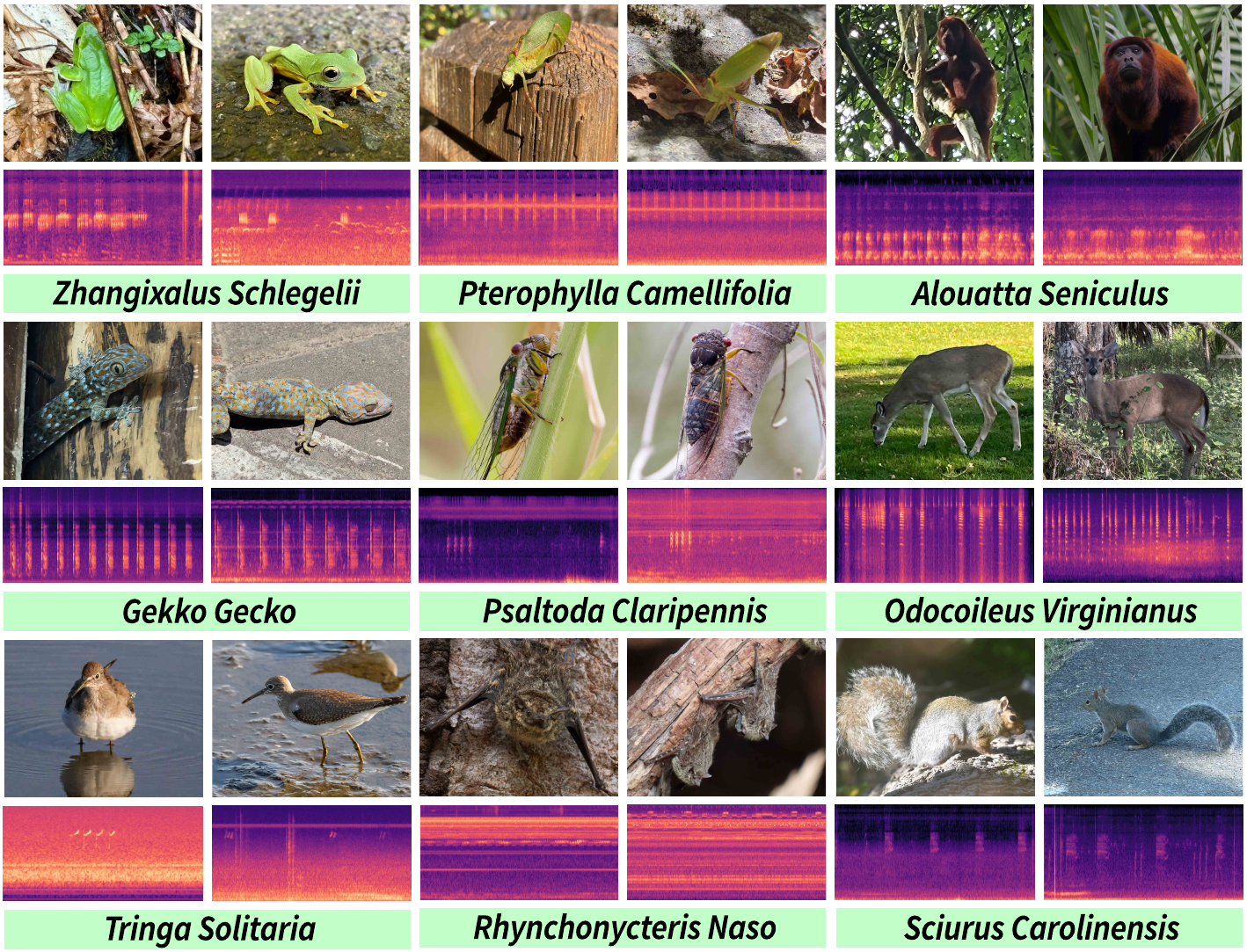}
\vspace{-14pt}
\caption{Examples from \scalebox{0.95}{$\mathsf{BioVITA\hspace{1.5pt}Train}$}.
Images and audio clips are shown with their corresponding scientific names.}
\label{fig:biovita_examples}
\vspace{-5pt}
\end{figure}

\paragraph{Examples.}
Figure~\ref{fig:biovita_examples} shows several examples from the constructed dataset.
As shown, when morphological differences among species are substantial, these distinctions become clearly visible in the mel-spectrogram visualizations, demonstrating the discriminative potential of acoustic representations.
Given the extensive diversity of species, \scalebox{0.95}{$\mathsf{BioVITA}$} introduces novel multimodal challenges.

\section{BioVITA Model}
\label{sec:model}
\begin{figure}
\centering
\includegraphics[width=\linewidth]{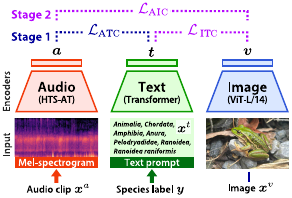}
\vspace{-18pt}
\caption{\scalebox{0.95}{$\mathsf{BioVITA\hspace{1.5pt}Model}$} consists of three encoders.
Building upon BioCLIP~2, we train the audio encoder in Stage 1, and jointly train the audio and text encoders in Stage 2.}
\label{fig:encoders}
\vspace{-5pt}
\end{figure}

This section presents \scalebox{0.95}{$\mathsf{BioVITA\hspace{1.5pt}Model}$}, a unified representation model.
As shown in Figure~\ref{fig:encoders}, our model consists of 
three encoders for the audio, image, and text modalities for taxonomy information.
To fully leverage well-established image-text encoders such as BioCLIP~2~\cite{Gu2025BioCLIP2}, we introduce a two-stage training framework that aligns audio representations to pre-trained image and text representations.
Due to inherent difficulties in distinguishing fine-grained visual and acoustic details, Stage~1 trains the audio encoder by minimizing only the audio-text contrastive (ATC) loss.

\subsection{Architectures}

\paragraph{Audio Encoder.} Following CLAP~\cite{laionclap2023}, we adopt HTS-AT \cite{Chen2022HTSAT} as the audio encoder, which is a hierarchical transformer consisting of four groups of SwinT~\cite{Liu2021SwinT} to extract audio representations from mel-spectrogram inputs.
The output dimension of the final projection layer is modified to obtain $d\!=\!768$ dimensional representations.
Given an input audio clip $\bm{x}^{a}$, we denote by $\bm{a}\!=\!f_{a}(\bm{x}^{a})\!\in\!\mathbb{R}^{d}$ the L2-normalized embedding extracted by the audio encoder $f_{a}$.

\paragraph{Image-Text Encoders.}
We adopt the pre-trained BioCLIP~2~\cite{Gu2025BioCLIP2}, which uses a ViT-L/14 as the image encoder and a 12-layer Transformer as the text encoder. Both of these encoders generate $768$-dimensional representations.
Given a text $\bm{t}$ and an image $\bm{v}$ as inputs, we denote by $\bm{t}\!=\!f_{t}(\bm{x}^{t})$ and $\bm{v}\!=\!f_{v}(\bm{x}^{v})$ the L2-normalized textual and visual representations, respectively, where $f_{t}$ is the text encoder and $f_{v}$ is the image encoder.

\subsection{Two-Stage Training}
\label{sec:stage1}

\paragraph{Stage\hspace{3pt}1\hspace{3pt}(Audio-Text).}
This stage aims to align audio and textual representations.
Let $\mathcal{B}\!=\!\{(\bm{x}^{a}_{i}, \bm{y}_{i})\}_{i=1}^{B}$ be a training mini-batch of size $B$, where each audio clip $\bm{x}^{a}_{i}$ is paired with its species label $\bm{y}_{i}$.
We first compute the audio-text similarity matrix $\bm{S}_{\scalebox{0.62}{\text{AT}}}\!\in\!\mathbb{R}^{B \times B}$ as $[S_{\scalebox{0.62}{\text{AT}}}]_{ij}\!=\!\bm{a}_{i}^{\top} \bm{t}_{j}/\tau$, where $\tau$ is the temperature hyperparameter.
Here, the text prompt $\bm{t}_{i}$ is generated from $\bm{y}_{i}$ by randomly selecting a pre-defined prompt template following BioCLIP~\cite{stevens2024bioclip}.
Subsequently, the ATC loss $\mathcal{L}_{\text{ATC}}$ is computed using row-wise and column-wise cross-entropy losses applied to the similarity matrix:
\begin{align}
\label{eq:loss_atc}
\mathcal{L}_{\text{ATC}}
=
\frac{1}{2}
\left(
\ell(\bm{S}_{\scalebox{0.62}{\text{AT}}}) +
\ell(\bm{S}_{\scalebox{0.62}{\text{AT}}}^{\top})
\right)
\end{align}
where $\ell(\bm{S})\!=\!- \frac{1}{B}\sum_{i=1}^{B} \log (\exp ([S]_{ii})/\sum_{j=1}^{B}\exp([S]_{ij}))$ is the cross-entropy loss. 
Training proceeds for 30 epochs using the AdamW optimizer with a constant learning rate of $10^{-4}$ and mini-batch size of 64.


\begin{figure*}
\centering
\includegraphics[width=\linewidth]{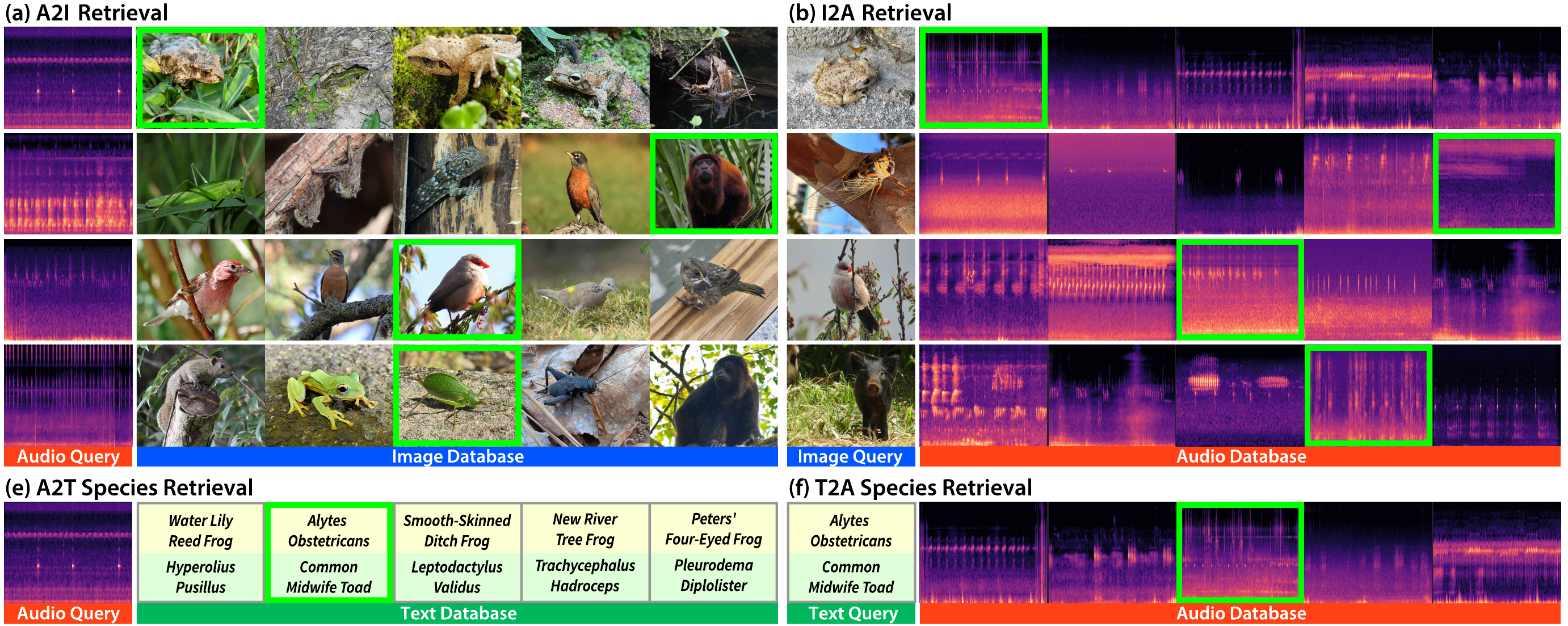}
\caption{
Task examples for \scalebox{0.95}{$\mathsf{BioVITA~Bench}$}.
Given a query, models are required to identify the relevant sample (indicated by green rectangles) from databases of 100 samples each. Five samples per database are shown.
}
\label{fig:benchmark_examples}
\end{figure*}

\paragraph{Stage\hspace{3pt}2\hspace{3pt}(VITA).}
After convergence of the ATC loss, we activate the AIC and ITC losses to achieve VITA alignment.
Given a training mini-batch
$\mathcal{B}\!=\!\{(\bm{x}^{a}_{i}, \bm{x}^{v}_{i}, \bm{y}_{i})\}_{i=1}^{B}$ consisting of audio-image-text triples,
Stage 2 minimizes the weighted sum of the three contrastive losses:
\begin{align}
\mathcal{L} = \mathcal{L}_{\text{ATC}}+\lambda (\mathcal{L}_{\text{AIC}} + \mathcal{L}_{\text{ITC}}),
\end{align}
where $\mathcal{L}_{\text{AIC}}$ and $\mathcal{L}_{\text{ITC}}$ are defined analogous to $\mathcal{L}_{\text{ATC}}$ in Eq.~\eqref{eq:loss_atc} using the audio-image similarity $[S_{\scalebox{0.62}{\text{AI}}}]_{ij}\!=\!\bm{a}_{i}^{\top} \bm{v}_{j}/\tau$ and the image-text similarity $[S_{\scalebox{0.62}{\text{IT}}}]_{ij}\!=\!\bm{v}_{i}^{\top} \bm{t}_{j}/\tau$, respectively.
Training continues for 10 epochs while halving the learning rate and making the audio and text encoders trainable.
To prevent an undesirable increase in the ATC loss minimized in Stage~1, we gradually increase the weight coefficient $\lambda$ from $0$ to $0.1$ using linear scheduling over the first 2 epochs. 

\paragraph{Setting.} We treat one epoch as at most 20 recordings per species.
To increase data diversity, we randomly crop each audio sample into 10-second segments.
For text prompts, we generate taxonomic descriptions and randomize their phrasing following the BioCLIP setup.

\section{BioVITA Benchmark}
\label{sec:benchmark}

This section presents \scalebox{0.95}{$\mathsf{BioVITA~Bench}$}, a novel benchmark for cross-modal species-level retrieval across image, text, and audio data.

\subsection{Benchmark Design}

We design fine-grained retrieval tasks to enable comprehensive analyses from multimodal, ecological, and generalization perspectives.

\paragraph{1)\hspace{2pt}Multimodal\hspace{2.2pt}Perspective.}\hspace{1pt}
To facilitate modality-specific analysis, we define six retrieval directions: image-to-audio (I2A), audio-to-image (A2I), image-to-text (I2T), text-to-image (T2I), audio-to-text (A2T), and text-to-audio (T2A) as shown in Figure~\ref{fig:benchmark_examples}.
These exhaustive directions systematically evaluate how effectively models handle multimodal biological data, while allowing comparison with bi-modal models using modality-specific subsets (\textit{e.g.}, A2T and T2A for CLAP).

\paragraph{2) Ecological Perspective.}\hspace{1pt}
For fine-grained ecological analysis, we define retrieval tasks at three taxonomic levels: Species, Genus, and Family.
This setup allows us to assess model performance not only at the species level but also at higher taxonomic levels, where category membership becomes broader.
Because visual and acoustic characteristics vary more widely within higher-level taxa, retrieval at the Family level represents a more challenging task.

\begin{table*}[t]
\centering
\small
\setlength{\tabcolsep}{4pt}
\caption{Species-level cross-modal retrieval results on the seen subset. We report Top 1 and 5 accuracies for each retrieval direction.}
\vspace{-2mm}
\label{tab:species_all_combined}
\begin{tabular}{lcccccccccccccc}
\toprule
\multirow{2}{*}{\textbf{Model}}
  & \multicolumn{2}{c}{\textbf{Audio$\rightarrow$Text}} 
  & \multicolumn{2}{c}{\textbf{Text$\rightarrow$Audio}} 
  & \multicolumn{2}{c}{\textbf{Audio$\rightarrow$Image}} 
  & \multicolumn{2}{c}{\textbf{Image$\rightarrow$Audio}} 
  & \multicolumn{2}{c}{\textbf{Image$\rightarrow$Text}} 
  & \multicolumn{2}{c}{\textbf{Text$\rightarrow$Image}}
  & \multicolumn{2}{c}{\textbf{Average}} \\[-0.2em]
\cmidrule(lr){2-3} \cmidrule(lr){4-5} \cmidrule(lr){6-7} \cmidrule(lr){8-9} \cmidrule(lr){10-11} \cmidrule(lr){12-13} \cmidrule(lr){14-15}
\\[-1.3em]
 & \scalebox{0.95}{\textbf{Top 1}} & \scalebox{0.95}{\textbf{Top 5}}
 & \scalebox{0.95}{\textbf{Top 1}} & \scalebox{0.95}{\textbf{Top 5}} & \scalebox{0.95}{\textbf{Top 1}} & \scalebox{0.95}{\textbf{Top 5}} & \scalebox{0.95}{\textbf{Top 1}} & \scalebox{0.95}{\textbf{Top 5}} & \scalebox{0.95}{\textbf{Top 1}} & \scalebox{0.95}{\textbf{Top 5}} & \scalebox{0.95}{\textbf{Top 1}} & \scalebox{0.95}{\textbf{Top 5}} & \scalebox{0.95}{\textbf{Top 1}} & \scalebox{0.95}{\textbf{Top 5}} \\[-0.2em]
\midrule
Random & 0.01 & 0.05 & 0.01 & 0.05 & 0.01 & 0.05 & 0.01 & 0.05 & 0.01 & 0.05 & 0.01 & 0.05 & 0.01 & 0.05\\
CLIP~\cite{clip}
& -- & --
& -- & --
& -- & --
& -- & --
& 50.9 & 65.1
& 63.8 & 72.5
& -- & -- \\
CLAP~\cite{laionclap2023}
  & 0.7 & 4.7
  & 1.0 & 4.7
  & -- & --
  & -- & --
  & -- & --
  & -- & --
  & -- & -- \\
ImageBind~\cite{girdhar2023imagebind}
   & 2.0 & 8.9 & 2.8 & 12.9 & 1.8 & 8.7 & 2.9 & 12.2 & 59.5 & 81.4 & 67.9 & 86.7 & 22.8 & 35.1 \\
BioCLIP~2~\cite{Gu2025BioCLIP2}
& -- & --
& -- & --
& -- & --
& -- & --
& 65.1 & 80.5 & 84.8 & 95.3
& -- & -- \\
TaxaBind~\cite{sastry2025taxabind}&9.6&28.1&11.8&32.9&13.3&35.8&16.2&41.4&56.9&80.0&66.9&86.9&29.1&50.9\\
\rowcolor{lightcyan} \scalebox{0.95}{$\mathsf{BioVITA}$} (Stage1)
   & 60.3 & 80.0 & 79.3 & 93.9 & 47.8 & 72.8 & 48.6 & 78.8 & 65.1 & 80.5 & 84.8 & 95.3 & 64.3 & 83.6 \\
\rowcolor{lightcyan} \scalebox{0.95}{$\mathsf{BioVITA}$} (Stage2)
   & 63.7 & 83.8 & 81.1 & 94.7 & 50.3 & 77.4 & 57.5 & 85.6 & 86.3 & 96.0 & 91.2 & 97.5 & 71.7 & 89.2 \\[-0.2em]

\bottomrule
\end{tabular}
\end{table*}

\begin{figure*}
\centering
\vspace{-2mm}
\includegraphics[width=\linewidth]{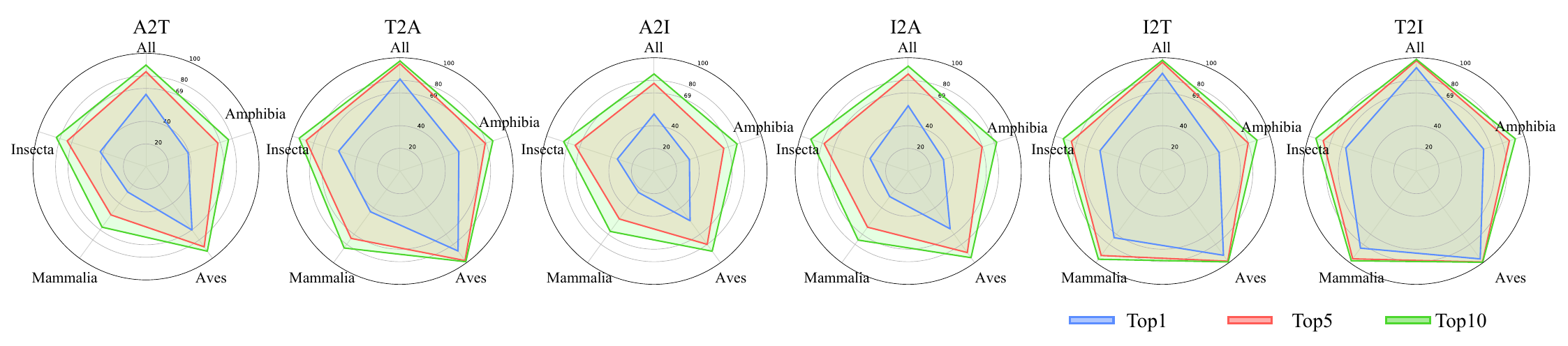}
\vspace{-7mm}
\caption{Accuracy by taxonomy class for each retrieval scenario.}
\label{fig:polar}

\end{figure*}

\begin{table}[t]
\centering 
\small
\caption{Prompt selection for \scalebox{0.95}{$\mathsf{BioVITA}$} on species-level retrieval.}
\vspace{-4pt}
\setlength{\tabcolsep}{2pt}
\begin{tabular}{lccccccc} 
\toprule
{Prompt} & \;{A2T}\; & \;{T2A}\; & \;{A2I}\; & \;{I2A}\; & \;{I2T}\; & \;{T2I}\; & \;{Avg.}\;\\
\midrule
Common Name \;\; &63.7&81.1&50.3&57.5&86.3&91.2&71.7\\
Science Name     &67.0&84.2&50.2&57.6&88.1&93.6&73.5\\
\bottomrule
\end{tabular}
\label{tb:prompt}
\vspace{-6mm}
\end{table}

\paragraph{3) Generalization Perspective.}\hspace{2pt}To evaluate generalizability, we categorize species into \textit{seen} and \textit{unseen} groups. 
Specifically, we create an unseen subset with species that are intentionally excluded from the training dataset.
This allows rigorous assessment of models' generalization abilities to previously unobserved taxa, closely reflecting realistic ecological scenarios where rare species may emerge.

\subsection{Task Definitions}

\paragraph{Scenarios and Tasks.}\hspace{2.5pt}By systematically combining the six modality directions, three ecological levels, and two generalization groups,
we obtain a total of 36 retrieval scenarios.
We define each scenario $\mathcal{S}$ as a set of $K$ independent retrieval tasks: $\mathcal{S}\!=\!\{(\bm{q}_{k}, \mathcal{D}_{k})\}_{k=1}^{K}$, where each task is represented by a pair of a query $\bm{q}_{k}$ and a database $\mathcal{D}_{k}$.
During evaluation, models perform retrieval for each task, identifying relevant samples from $\mathcal{D}_{k}$ given each query $\bm{q}_{k}$.

\paragraph{Queries and Databases.}
Each query $\bm{q}_{k}$ is presented in one modality (image, text, or audio), while the database $\mathcal{D}_{k}$ exclusively contains samples from one of the two remaining modalities.
For example, in A2I retrieval, $\bm{q}_{k}$ is an audio clip and $\mathcal{D}_{k}$ is a set of images.
Each database contains $K=100$ samples corresponding to the specified level and generalization subset.
One of these samples is directly relevant to $\bm{q}_{k}$, serving as the positive target, while the remaining 99 samples act as distractors.

\paragraph{High-Level Retrieval.}
We also construct retrieval tasks at the genus and family levels.
For these tasks, each query and its candidates are drawn from different species within the same genus or family, while keeping the retrieval setting fixed to 100-way retrieval across all taxonomic levels.
Because species within a family can be visually and acoustically diverse, these higher-level tasks are more challenging than species-level retrieval.

\paragraph{Database Construction.}
The databases are constructed via random sampling from test sets of each modality. The audio test set consists of the audio clips reserved in Sec.~\ref{sec:dataset_construction}. For the image test set, we curate a new collection of 128,645 images from iNaturalist, ensuring it is disjoint from the ToL-200M dataset. 

\paragraph{Evaluation Metrics.}
Top-1 and Top-5 accuracies are used as evaluation metrics on each retrieval scenario. We also report average accuracy.

\begin{table*}[t]
\centering
\small
\setlength{\tabcolsep}{3pt}
\caption{
High-level retrieval results at the Genus and Family levels, where queries and targets belong to different species within the same taxon. This setting is more challenging as it requires models to capture semantic relationships beyond exact species-level identification.}
\vspace{-5pt}
\label{tab:genus_family_all_combined}
\begin{tabular}{clcccccccccccccc}
\toprule
\multicolumn{2}{l}{\multirow{2}{*}{\textbf{Model}}}
  & \multicolumn{2}{c}{\textbf{Audio$\rightarrow$Text}} 
  & \multicolumn{2}{c}{\textbf{Text$\rightarrow$Audio}} 
  & \multicolumn{2}{c}{\textbf{Audio$\rightarrow$Image}} 
  & \multicolumn{2}{c}{\textbf{Image$\rightarrow$Audio}} 
  & \multicolumn{2}{c}{\textbf{Image$\rightarrow$Text}} 
  & \multicolumn{2}{c}{\textbf{Text$\rightarrow$Image}}
  & \multicolumn{2}{c}{\textbf{Average}} \\[-0.2em]
\cmidrule(lr){3-4} \cmidrule(lr){5-6} \cmidrule(lr){7-8} \cmidrule(lr){9-10} \cmidrule(lr){11-12} \cmidrule(lr){13-14} \cmidrule(lr){15-16}
\\[-1.3em]
&  & \scalebox{0.95}{\textbf{Top 1}} & \scalebox{0.95}{\textbf{Top 5}}
 & \scalebox{0.95}{\textbf{Top 1}} & \scalebox{0.95}{\textbf{Top 5}} & \scalebox{0.95}{\textbf{Top 1}} & \scalebox{0.95}{\textbf{Top 5}} & \scalebox{0.95}{\textbf{Top 1}} & \scalebox{0.95}{\textbf{Top 5}} & \scalebox{0.95}{\textbf{Top 1}} & \scalebox{0.95}{\textbf{Top 5}} & \scalebox{0.95}{\textbf{Top 1}} & \scalebox{0.95}{\textbf{Top 5}} & \scalebox{0.95}{\textbf{Top 1}} & \scalebox{0.95}{\textbf{Top 5}} \\[-0.2em]
\midrule
\multirow{5}{*}{\rotatebox{90}{\textbf{Genus}}}
& ImageBind~\cite{girdhar2023imagebind}
& 1.3 & 6.1 & 1.6 & 8.3 & 1.8 & 9.0 & 2.7 & 12.1 & 6.6 & 14.2 & 7.3 & 14.0 & 3.6 & 10.6 \\
& BioCLIP~2~\cite{Gu2025BioCLIP2}
& -- & -- & -- & -- & -- & -- & -- & -- & 19.3 & 34.5 & 71.7 & 87.0  & -- & -- \\
& TaxaBind~\cite{sastry2025taxabind}&5.9&18.4&14.1&34.2&11.7&33.8&14.9&54.3&30.8&51.7&50.5&72.3&21.3&44.1\\
& \cellcolor{lightcyan}\scalebox{0.95}{$\mathsf{BioVITA}$} (Stage1)
& \cellcolor{lightcyan}37.3 & \cellcolor{lightcyan}58.1 & \cellcolor{lightcyan}80.9 & \cellcolor{lightcyan}91.1 & \cellcolor{lightcyan}41.8 & \cellcolor{lightcyan}67.9 & \cellcolor{lightcyan}42.8 & \cellcolor{lightcyan}72.5 & \cellcolor{lightcyan}19.3 & \cellcolor{lightcyan}34.5 & \cellcolor{lightcyan}71.7 & \cellcolor{lightcyan}87.0 & \cellcolor{lightcyan}49.0 & \cellcolor{lightcyan}68.5 \\
& \cellcolor{lightcyan}\scalebox{0.95}{$\mathsf{BioVITA}$} (Stage2)
& \cellcolor{lightcyan}\textbf{53.6} & \cellcolor{lightcyan}\textbf{75.4} & \cellcolor{lightcyan}\textbf{84.9} & \cellcolor{lightcyan}\textbf{94.2} & \cellcolor{lightcyan}\textbf{43.2} & \cellcolor{lightcyan}\textbf{71.5} & \cellcolor{lightcyan}\textbf{48.6} & \cellcolor{lightcyan}\textbf{78.0} & \cellcolor{lightcyan}\textbf{74.7} & \cellcolor{lightcyan}\textbf{89.6} & \cellcolor{lightcyan}\textbf{90.7} & \cellcolor{lightcyan}\textbf{96.1} & \cellcolor{lightcyan}\textbf{66.0} & \cellcolor{lightcyan}\textbf{84.1} \\[-0.2em]
\midrule
\multirow{5}{*}{\rotatebox{90}{\textbf{Family}}}
& ImageBind~\cite{girdhar2023imagebind}
& 1.1 & 5.3 & 1.3 & 7.5 & 2.0 & 8.7 & 2.5 & 11.0 & 6.9 & 14.5 & 4.7 & 13.1 & 3.1 & 10.0 \\
& BioCLIP~2~\cite{Gu2025BioCLIP2}
& -- & -- & -- & -- & -- & -- & -- & -- & 11.9 & 25.5 & 35.8 & 58.9 & -- & -- \\
& TaxaBind~\cite{sastry2025taxabind}&3.6&11.7&7.0&21.2&9.6&27.5&12.9&31.9&11.5&24.0&22.5&37.5&11.2&25.6\\
& \cellcolor{lightcyan}\scalebox{0.95}{$\mathsf{BioVITA}$} (Stage1)
& \cellcolor{lightcyan}17.2 & \cellcolor{lightcyan}35.5 & \cellcolor{lightcyan}34.2 & \cellcolor{lightcyan}55.7 & \cellcolor{lightcyan}31.8 & \cellcolor{lightcyan}56.2 & \cellcolor{lightcyan}31.7 & \cellcolor{lightcyan}57.2 & \cellcolor{lightcyan}11.9 & \cellcolor{lightcyan}25.5 & \cellcolor{lightcyan}35.8 & \cellcolor{lightcyan}58.9 & \cellcolor{lightcyan}27.1 & \cellcolor{lightcyan}48.2 \\
& \cellcolor{lightcyan}\scalebox{0.95}{$\mathsf{BioVITA}$} (Stage2)
& \cellcolor{lightcyan}\textbf{19.2} & \cellcolor{lightcyan}\textbf{38.2} & \cellcolor{lightcyan}\textbf{37.1} & \cellcolor{lightcyan}\textbf{55.0} & \cellcolor{lightcyan}\textbf{32.8} & \cellcolor{lightcyan}\textbf{60.0} & \cellcolor{lightcyan}\textbf{38.8} & \cellcolor{lightcyan}\textbf{66.1} & \cellcolor{lightcyan}\textbf{36.9} & \cellcolor{lightcyan}\textbf{58.0} & \cellcolor{lightcyan}\textbf{53.3} & \cellcolor{lightcyan}\textbf{74.5} & \cellcolor{lightcyan}\textbf{36.4} & \cellcolor{lightcyan}\textbf{58.6} \\[-0.2em]
\bottomrule
\end{tabular}
\end{table*}

\begin{table*}[t]
\centering
\small
\setlength{\tabcolsep}{4pt}
\caption{Species-level cross-modal retrieval results on the unseen subset. We report Top 1 and 5 accuracies for each retrieval direction.}
\vspace{-5pt}
\label{tab:unseen}
\begin{tabular}{lcccccccccccccc}
\toprule
\multirow{2}{*}{\textbf{Model}}
  & \multicolumn{2}{c}{\textbf{Audio$\rightarrow$Text}} 
  & \multicolumn{2}{c}{\textbf{Text$\rightarrow$Audio}} 
  & \multicolumn{2}{c}{\textbf{Audio$\rightarrow$Image}} 
  & \multicolumn{2}{c}{\textbf{Image$\rightarrow$Audio}} 
  & \multicolumn{2}{c}{\textbf{Image$\rightarrow$Text}} 
  & \multicolumn{2}{c}{\textbf{Text$\rightarrow$Image}}
  & \multicolumn{2}{c}{\textbf{Average}} \\[-0.2em]
\cmidrule(lr){2-3} \cmidrule(lr){4-5} \cmidrule(lr){6-7} \cmidrule(lr){8-9} \cmidrule(lr){10-11} \cmidrule(lr){12-13} \cmidrule(lr){14-15}
\\[-1.3em]
 & \scalebox{0.95}{\textbf{Top 1}} & \scalebox{0.95}{\textbf{Top 5}}
 & \scalebox{0.95}{\textbf{Top 1}} & \scalebox{0.95}{\textbf{Top 5}} & \scalebox{0.95}{\textbf{Top 1}} & \scalebox{0.95}{\textbf{Top 5}} & \scalebox{0.95}{\textbf{Top 1}} & \scalebox{0.95}{\textbf{Top 5}} & \scalebox{0.95}{\textbf{Top 1}} & \scalebox{0.95}{\textbf{Top 5}} & \scalebox{0.95}{\textbf{Top 1}} & \scalebox{0.95}{\textbf{Top 5}} & \scalebox{0.95}{\textbf{Top 1}} & \scalebox{0.95}{\textbf{Top 5}} \\[-0.2em]
\midrule
ImageBind~\cite{girdhar2023imagebind} & 2.2 & 11.8 & 1.5 & 8.9 & 1.1 & 9.2 & 3.0 & 12.1 & 42.8 & 67.5 & 47.6 & 71.5 & 16.4 & 30.2 \\
BioCLIP~2~\cite{Gu2025BioCLIP2}
& -- & -- & -- & -- & -- & -- & -- & -- &33.0 & 59.5 & 37.1 & 66.7&--&--\\
\rowcolor{lightcyan} \scalebox{0.95}{$\mathsf{BioVITA}$} (Stage1)
  & 14.6 & 32.9 & 20.0 & 43.4 & 7.8 & 23.6 & 6.4 & 25.8 & 33.0 & 59.5 & 37.1 & 66.7 & 19.8 & 42.0 \\
\rowcolor{lightcyan} \scalebox{0.95}{$\mathsf{BioVITA}$} (Stage2)
   & 36.6 & 62.2 & 54.1 & 74.8 & 29.3 & 54.3 & 28.0 & 55.5 & 76.8 & 94.7 & 86.5 & 96.2 & 51.9 & 73.0 \\[-0.2em]
\bottomrule
\end{tabular}
\end{table*}

\section{Experiments}
\label{sec:experiments}

We conduct extensive experiments to evaluate the BioVITA framework.

\subsection{Species-Level Cross-Modal Retrieval}
\label{sec:exp_seen}

\paragraph{Settings.} We perform cross-modal retrieval at the species level and analyze the results from a multimodal perspective.
To demonstrate the effectiveness of \scalebox{0.95}{$\mathsf{BioVITA~Model}$}, we implement four state-of-the-art baselines: CLIP~\cite{clip}, CLAP~\cite{laionclap2023}, ImageBind~\cite{girdhar2023imagebind}, 
BioCLIP~2~\cite{Gu2025BioCLIP2}, and TaxaBind~\cite{sastry2025taxabind}.
Among these, CLIP and BioCLIP~2 support image-text modalities, CLAP supports audio-text modalities, while ImageBind integrates all three modalities.
We utilize official implementations and pretrained checkpoints for all baseline models, employing cosine similarity between representations to measure cross-modal similarity during retrieval.\looseness=-1

\paragraph{Results.}
Table~\ref{tab:species_all_combined} summarizes results across the six cross-modal directions.
Our \scalebox{0.95}{$\mathsf{BioVITA}$} effectively handles all retrieval scenarios and significantly outperforms the tri-modal baseline (ImageBind), achieving average Top-1 and Top-5 accuracies of 71.7\% and 89.2\%, respectively.
Stage~1 training (audio-text alignment) alone already achieves substantial gains, demonstrating the benefit of grounding audio features with BioCLIP~2 via the ATC~(audio-text contrastive)loss.
Stage~2, which incorporates visual information, further improves all retrieval scenarios by providing complementary cues for robust VITA alignment.
We also observe performance improvements in image-text retrieval tasks over BioCLIP~2 at Stage~2, indicating that VITA alignment enriches image-text representations.

Table~\ref{tb:prompt} shows the differences in text prompt settings used in \scalebox{0.95}{$\mathsf{BioVITA}$}. For inference, we use the common name for the retrieval target in Table~\ref{tab:species_all_combined} to ensure a fair comparisons, as general-purpose models are typically trained on data that use culturally assigned common names. Meanwhile, when using scientific names in the prompt, we observe higher accuracy. 
This suggests that scientific names provide clearer taxonomic information than culturally assigned common names, which enables the model to utilize the hierarchical taxonomic structure, learned during the training phase, more effectively. 

\paragraph{Performance by Taxonomy Class.}
Figure~\ref{fig:polar} analyzes the accuracy by taxonomy class. In audio-related tasks, the highest accuracy is observed for \textit{Aves} (birds), followed by \textit{Insecta} (insects), \textit{Amphibia} (amphibians), and \textit{Mammalia} (mammals). 
Birds typically produce species-specific vocalizations with distinctive acoustic patterns, enabling accurate identification. Moreover, the frequency with which birds are acoustically observed has led to the availability of rich training data.
In contrast, mammalian vocalizations vary significantly with body size and are more difficult to distinguish from ambient noise.

\begin{figure}
\centering
\includegraphics[width=\linewidth]{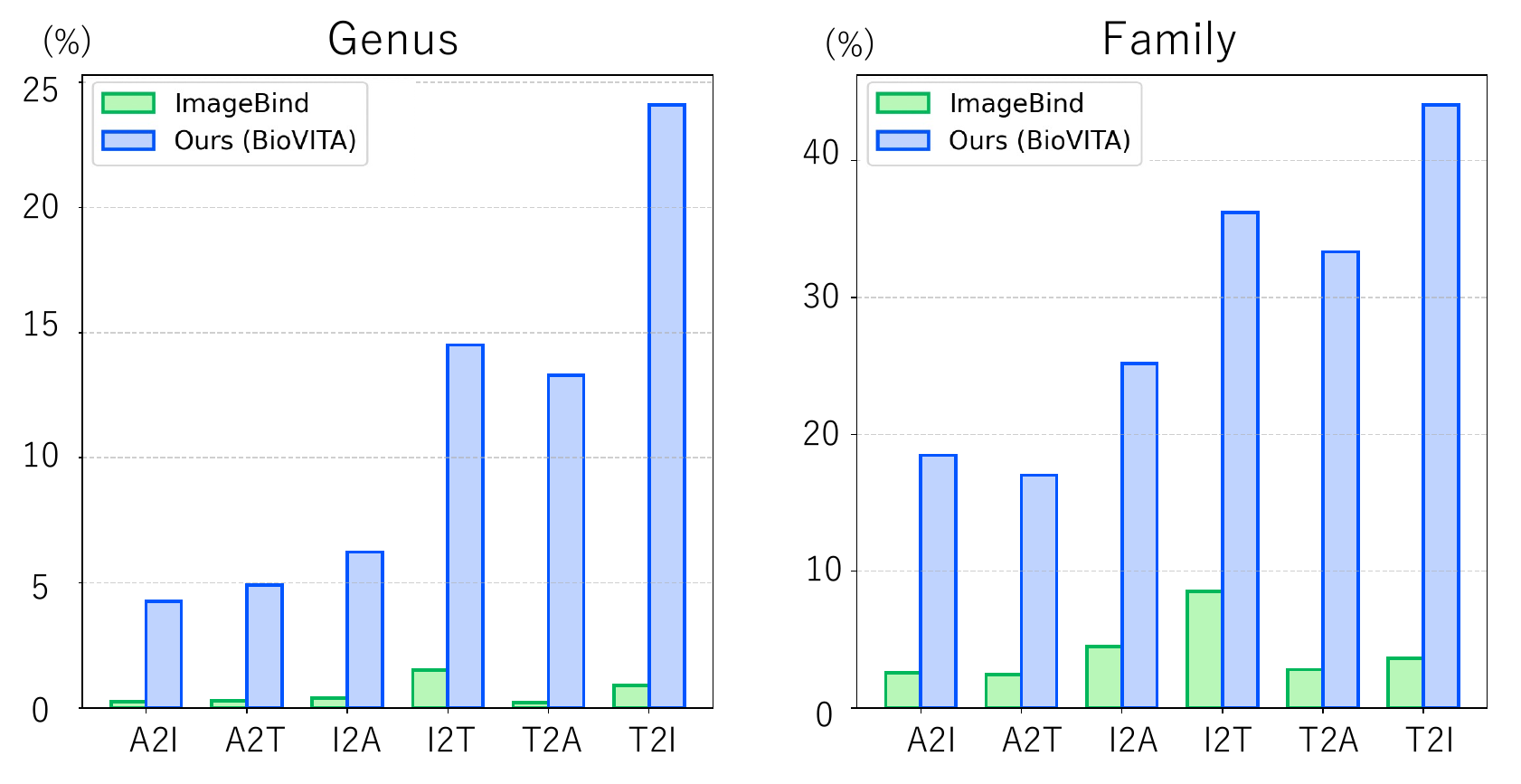}
\vspace{-14pt}
\caption{Genus and family-level consistency of Top-1 misclassifications across retrieval tasks. The \scalebox{0.95}{$\mathsf{BioVITA}$} model more frequently predicts the correct genus or family despite species-level errors.}
\label{fig:qualitative}
\vspace{-9pt}
\end{figure}

\subsection{High-Level Retrieval}
We further evaluate retrieval performance at higher taxonomic levels, specifically at the genus and family levels, where a retrieval is considered correct if the query and target samples share the same genus or family, regardless of their species identity.
This setting increases the difficulty: models can no longer rely solely on exact species-level matches and must instead capture broader ecological and taxonomic similarities among related organisms.\looseness=-1

\paragraph{Results.}
Table~\ref{tab:genus_family_all_combined} reports high-level retrieval performance comparing \scalebox{0.95}{$\mathsf{BioVITA}$} with ImageBind and BioCLIP 2.
Overall, \scalebox{0.95}{$\mathsf{BioVITA}$} (Stage~2) achieves superior performance across all retrieval directions and levels.
From Stage~1 to Stage~2, we observe accuracy improvements consistent with species-level retrieval trends. Improvements are particularly pronounced in audio-image retrieval scenarios (A2I, I2A).

Meanwhile, the retrieval performance itself decreases at higher taxonomic levels compared to the species level.
This might arise from the intrinsic diversity within families, as species belonging to the same family can differ significantly in both their visual appearance and acoustic properties.
Consequently, “retrieving the correct family” does not correspond to a tightly clustered embedding region, making the task inherently challenging.

Importantly, this decrease does not imply that the model fails to learn hierarchical structure.
Our taxonomy-aware prompting, inspired by BioCLIP, helps the model encode meaningful hierarchical relationships.
Figure \ref{fig:qualitative} shows our species-level retrieval error analysis.
The left plot reports, among all species-level errors, the proportion in which the retrieved sample belongs to the correct genus, and the right plot reports the proportion belonging to the correct family.
These results indicate that although species-level predictions are sometimes incorrect, our method is more likely than ImageBind to retrieve samples that match the query at higher taxonomic levels. This suggests that the learned representations successfully capture hierarchical taxonomic structure.

\subsection{Can BioVITA Generalize to Unseen Species?}
\label{sec:exp_unseen}

To investigate the generalization capabilities of \scalebox{0.95}{$\mathsf{BioVITA}$}, we evaluate its retrieval performance on the unseen subset, comprising 325 species intentionally withheld during our training.
The settings are identical to those in Sec.~\ref{sec:exp_seen}.\looseness=-1

\paragraph{Results.}
Table~\ref{tab:unseen} summarizes the retrieval results. Despite encountering entirely novel taxa, \scalebox{0.95}{$\mathsf{BioVITA}$} demonstrates robust generalization, achieving average Top-1 and Top-5 accuracies of 51.9\% and 73.0\%, respectively.
Consistent with observations in seen scenarios, the improvement from Stage~1 to Stage~2 underscores the crucial role of incorporating visual modalities for enhancing generalization.

{\renewcommand{\arraystretch}{0.85}
\begin{table}[t]
\centering
\caption{Ecological trait prediction for unseen species. All models are fine-tuned on the proposed training dataset and evaluated using the F1 score.}
\label{tab:traits}
\vspace{-8pt}
\centering
\small
\setlength{\tabcolsep}{6pt}
\begin{tabular}{p{0mm}lcccc}
\toprule
\multicolumn{2}{l}{\multirow{2}{*}{\textbf{Category}}} & \multirow{2}{*}{CLAP} & \multirow{2}{*}{CLIP} & \scalebox{0.95}{$\mathsf{BioVITA}$} & \scalebox{0.95}{$\mathsf{BioVITA}$} \\[-0.1em]
&&&&(Audio)&(Image)\\
\midrule
\multicolumn{2}{l}{Diet Type} & 43.7 & 41.5 & 53.9 & 41.1 \\

\multicolumn{2}{l}{Act. Pattern} & 80.9 &82.1  & 83.7 &  82.2\\

\multicolumn{2}{l}{Locomotion} & 79.7 & 84.3 & 81.1 &  83.1\\

\multicolumn{2}{l}{Lifestyle} &  82.7 & 92.5 &85.1 & 89.9 \\

\multicolumn{2}{l}{Trophic Role} & 79.4 & 89.2 & 89.7 & 88.0 \\

\multicolumn{2}{l}{Habitat} & 62.4 & 58.7 & 64.9 & 64.9 \\

\multicolumn{2}{l}{Clim. Dist.} &  76.3 & 74.4 &  78.7 &75.4  \\

\multicolumn{2}{l}{Social Behavior} & 76.7 & 78.6 & 77.6 & 79.3 \\

\multicolumn{2}{l}{Migration} & 22.4 & 0.0 & 45.7 & 10.6 \\
\bottomrule
\end{tabular}
\vspace{-3mm}
\end{table}}

\begin{table}[t]
\centering 
\small
\caption{Ablation study. We report Top-1 accuracy for each retrieval direction.}
\vspace{-8pt}
\setlength{\tabcolsep}{1pt}
\begin{tabular}{lccccccc} 
\toprule
{Method} & {A2T} & {T2A} & {A2I} & {I2A} & {I2T} & {T2I} & {Avg.}\\
\midrule
\rowcolor{lightcyan} \scalebox{0.95}{$\mathsf{BioVITA}$} (full) & 63.7&81.1&50.3&57.5&86.3&91.2&71.7\\
\midrule
\multicolumn{8}{l}{\textit{Stage}}\\
w/o Stage1&20.7&29.6&10.3&15.1&49.9&45.6&34.2\\
w/o Stage2 &60.3&79.3&47.8&48.6&65.1&84.8&64.3\\
\midrule
\multicolumn{8}{l}{\textit{Stage1 Initial Weight}}\\
w/o BioCLIP2&15.8&18.2&1.32&1.17&50.9&63.8&25.1\\
w BioCLIP2&60.3&79.3&47.8&48.6&65.1&84.8&64.3\\
\midrule
\multicolumn{8}{l}{\textit{Training Strategy}}\\
w/o Text-Encoder learning\;\;&63.0&73.3&50.1&50.1&65.1&84.8&64.5\\
w/o $\mathcal{L}_{\text{ITC}}$&60.9&78.3&50.0&57.9&54.3&87.7&64.9\\
w/o $\mathcal{L}_{\text{AIC}}$&62.1&80.9&40.0&27.7&85.5&90.8&64.5\\
\bottomrule
\end{tabular}
\vspace{-3mm}
\label{tb:ablation}
\end{table}

\subsection{Trait Prediction}
We further evaluate \scalebox{0.95}{$\mathsf{BioVITA}$} on ecological trait prediction.
Table~\ref{tab:traits} summarizes results across various trait categories.
All models are trained with an additional linear layer on top of the encoder.
For CLAP and \scalebox{0.95}{$\mathsf{BioVITA}$} (Audio), the models are trained on ecological trait representations and acoustic data, whereas CLIP and \scalebox{0.95}{$\mathsf{BioVITA}$} (Image) are trained on image features and their associated traits.

\scalebox{0.95}{$\mathsf{BioVITA}$} checkpoints enable models to learn ecological traits efficiently, particularly in the audio modality.
In the audio modality, the performance gain is especially pronounced for behavioral traits such as trohabitat and migration, suggesting that acoustic representations are particularly effective at capturing the temporal and behavioral characteristics inherent in vocalizations.
For example, urban birds often shift their song frequencies to compensate for anthropogenic noise, and vegetation structure can shape acoustic signals through habitat-specific adaptations~\cite{job2016song}.
Such properties allow acoustic modalities to effectively encode behavioral traits, which explains the observed improvements.

\subsection{Ablation Study}
We conduct an ablation study to analyze the impact of each component within \scalebox{0.95}{$\mathsf{BioVITA}$}.
As shown in Table~\ref{tb:ablation}, removing Stage~1 degrades performance, indicating that this stage is essential for guiding the initial alignment.
Training from scratch is also suboptimal, underscoring the importance of leveraging pre-trained visual-textual representations from BioCLIP~2.
These results validate that each training component contributes meaningfully to robust VITA alignment.

\section{Conclusion}
We introduced \scalebox{0.95}{$\mathsf{BioVITA}$}, a comprehensive framework for biological VITA alignment.
With a large-scale tri-modal dataset encompassing audio clips and images from 14k species, we proposed a two-stage training pipeline to effectively unify representations.
Our experiments demonstrated superior performance across diverse retrieval scenarios and ecological trait predictions, highlighting \scalebox{0.95}{$\mathsf{BioVITA}$}'s ability to capture nuanced behavioral and ecological signals.

\section{Acknowledgments}
This work was partly supported by JSPS KAKENHI JP25K24368, JST FOREST JPMJFR206F, and JST ASPIRE JPMJAP2502.

{
    \small
    \bibliographystyle{ieeenat_fullname}
    \bibliography{main}
}
\clearpage


\appendix
\maketitlesupplementary

\noindent In this supplementary material, we provide
\begin{description}[labelwidth=1.0em, labelsep=0.5em, leftmargin=1.5em]
  \setlength{\parskip}{0cm}
  \setlength{\itemsep}{0cm} 
\item[A.] implementation details of baseline models and prompts,
\item[B.] details on the process of collecting the image dataset,
\item[C.] detailed analyses including additional comparisons, visualizations, and training data size, and
\item[D.] more dataset details, including visualizations of examples and distributions.
\end{description}

\section{Implementation Details}
\subsection{Model Details}

\begin{itemize}
    \item \textbf{CLIP}~\cite{clip}: CLIP is a large-scale image-text contrastive model trained on diverse web data.  It projects images and texts into a shared embedding space using a ViT-based image encoder and a Transformer-based text encoder. As a unimodal-pair model, it supports only image--text alignment but provides a strong vision-language baseline for cross-modal retrieval tasks.
    
    \item \textbf{CLAP}~\cite{laionclap2023}: CLAP extends the CLIP framework to audio-text modalities by introducing an audio encoder trained jointly with a text encoder. 
    It enables zero-shot recognition and retrieval across audio and text domains. 
    Since CLAP does not model images, it serves as the primary baseline for evaluating audio-text alignment in biological settings.\looseness=-1

    \item \textbf{ImageBind}~\cite{girdhar2023imagebind}: ImageBind is a unified multi-modal model that binds six different modalities, including audio, image, and text, into a single representation space. 
    It leverages image embeddings as the central hub, learning cross-modal correspondences through large-scale contrastive training. 
    As a tri-modal model, ImageBind provides a comprehensive reference point for evaluating unified audio–image–text alignment.

    \item \textbf{BioCLIP~2}~\cite{Gu2025BioCLIP2}: BioCLIP~2 is a biology-specialized vision-language model based on ViT-L/14 for images and a 12-layer Transformer for text. 
    Trained on large-scale curated biological datasets, it achieves strong fine-grained species-level discrimination. 
    In our evaluation, BioCLIP~2 serves both as a strong image-text baseline and as the vision-language foundation for our \scalebox{0.95}{$\mathsf{BioVITA}$} model.\looseness=-1
\end{itemize}

\subsection{Prompt Details}
To incorporate taxonomy information into audio-text embeddings, we train the CLAP model with prompts augmented by 1) Common name (Com), 2) Scientific name (Sci), and 3) Taxonomic sequence (Tax) following the BioCLIP~2 setting.
The augmentation function $\phi$ randomly selects one of the five prompts (Com, Sci, Tax, Sci+Com, and Tax+Com) defined in Table~\ref{tab:tax_ex}.
For instance, given the common species' name (\eg, {`Anianiau}),
the augmented prompts include the scientific name (\eg, \textit{Magumma parva}) and its taxonomic order (\eg, \textit{Aves Passeriforme}).

\subsection{Model Parameter Sizes}
Our tri-modal model {$\mathsf{BioVITA}$} consists of three encoders: a BioCLIP2 image tower,
a BioCLIP2 text tower, and a CLAP audio encoder, together with a linear
audio-to-vision/text projection layer.  
We report the exact parameter counts obtained from the instantiated
PyTorch model.

\begin{itemize}
    \item BioCLIP2 image encoder (visual tower):
    303.97M parameters.  
    All weights are frozen during training.

    \item BioCLIP2 text encoder:
    123.65M parameters.  
    We freeze the entire transformer stack; only 
    0.65M parameters remain trainable.

    \item CLAP audio encoder:
    153.49M parameters.  
    All parameters are trainable.

    \item Audio projection layer:
    A linear adapter with 0.39M parameters.
\end{itemize}

In total, the model contains 581.5 M parameters
, of which 154.5M parameters are trainable.

We trained our model on 8$\times$V100 GPUs (32\,GB each). 
Because the dataset is stored on a separate storage server, 
the data transfer overhead increases the overall training time.
Stage~1 training required approximately two days, 
while Stage~2 took about one day.

\begin{table}[t]
\centering
\small
\resizebox{0.5\textwidth}{!}{
\begin{tabular}{llp{0.4\textwidth}}
\hline
\textbf{Template}  & \textbf{Example (\textquoteleft  Anianiau)} \\
\hline
Common Name (Com) & \textquoteleft Anianiau \\
Scientific Name (Sci) & Magumma Parva \\
Taxonomic Sequence (Tax) & Aves Passeriformes, Fringillidae Magumma, Magumma Parva \\
Sci + Com & Magumma Parva with a common name \textquoteleft Anianiau \\
\multirow{2}{*}{Tax + Com} & Aves Passeriformes, Fringillidae Magumma, Magumma Parva, \\
& with a common name \textquoteleft Anianiau \\
\hline
\end{tabular}
}
\caption{Examples of textual descriptions following the five templates used in training.}
\label{tab:tax_ex}
\end{table}

\section{Test Dataset Creation}
\subsection{Audio}
After collecting the audio data for 14K species, we split it into training and test sets with a 9:1 ratio, holding out 325 species that remain completely unseen during training. This results in approximately 1.3M training samples and 44K test samples. For efficient evaluation, we limit the number of audio clips per species to around ten.

For benchmarking, we construct 100-option multiple-choice questions, yielding roughly 30K species-level questions, 10K genus-level questions, and 1.7K family-level questions for each task type. The dataset covers a total of 9,725 species.

\subsection{Image}
To ensure fairness in building our model, we carefully curate the training and test data for the image modality while preventing any leakage.

For training, we collect images from ToL-200M for all species that appear in the audio training set. 
This results in 12,916 species, covering 91.4\% of the species in the audio training split.

For testing, we collect images from iNaturalist using the iNaturalist API by querying species names that appear in the audio test set.  
Since ToL-200M provides the original image source URLs, we can extract the corresponding iNaturalist observation IDs directly from these URLs.  
During test image collection, we exclude all images whose observation IDs match those extracted from ToL-200M, ensuring that no images overlap between the training and test splits.   
We also apply GroundingDINO~\cite{gdino} to filter images: if no animal is detected when using "animal" as the text prompt, we exclude the corresponding image.
We additionally verify within each species that the retrieved test images are distinct from the training set. Finally, we obtain a clean set of 128,645 images from 9,487 species that do not overlap with the ToL-200M dataset.

\section{Further Analysis}
\subsection{Model Comparison}
\begin{table*}[t]
\centering
\small
\setlength{\tabcolsep}{3pt}
\caption{Classification results of BioLingual and \scalebox{0.95}{$\mathsf{BioVITA}$}. Since BioLingual may include part of our test data in its training split, we construct a new 2024 audio test set and evaluate the classification performance on it.}
\label{tab:taxon_species_family_genus}
\begin{tabular}{llcccccccccccccccc}
\toprule
\multirow{3}{*}{\textbf{Model}}
&
  & \multicolumn{4}{c}{\textbf{Species}}
  & \multicolumn{4}{c}{\textbf{Genus}}   
  & \multicolumn{4}{c}{\textbf{Family}}  
  & \multicolumn{4}{c}{\textbf{Average}} \\[-0.2em]

\cmidrule(lr){3-6}
\cmidrule(lr){7-10}
\cmidrule(lr){11-14}
\cmidrule(lr){15-18}
\\[-1.3em]

& 
& \multicolumn{2}{c}{Audio$\rightarrow$Text} 
& \multicolumn{2}{c}{Text$\rightarrow$Audio}
& \multicolumn{2}{c}{Audio$\rightarrow$Text} 
& \multicolumn{2}{c}{Text$\rightarrow$Audio}
& \multicolumn{2}{c}{Audio$\rightarrow$Text} 
& \multicolumn{2}{c}{Text$\rightarrow$Audio}
& \multicolumn{2}{c}{Audio$\rightarrow$Text}
& \multicolumn{2}{c}{Text$\rightarrow$Audio} \\[-0.2em]
\cmidrule(lr){3-4} \cmidrule(lr){5-6} \cmidrule(lr){7-8} \cmidrule(lr){9-10} \cmidrule(lr){11-12} \cmidrule(lr){13-14} \cmidrule(lr){15-16}\cmidrule(lr){17-18}
\\[-1.3em]
& 
& Top1 & Top5 & Top1 & Top5
& Top1 & Top5 & Top1 & Top5
& Top1 & Top5 & Top1 & Top5
& Top1 & Top5 & Top1 & Top5 \\
\midrule

BioLingual~\cite{bioacoustics}
&  & 24.0 & 45.9 & 26.5 & 50.3   
& 23.2 & 42.8 & 51.7 & 68.2      
& 5.4  & 15.9 & 20.2 & 38.9      
& 17.5 & 34.9 & 32.8 & 52.5 \\   

\rowcolor{lightcyan}\scalebox{0.95}{$\mathsf{BioVITA}$} 
& & 24.4 & 49.2 & 27.7 & 56.6   
& 32.5 & 51.4 & 58.0 & 74.7      
& 17.4 & 42.9 & 52.3 & 77.3      
& 24.8 & 47.9 & 46.0 & 69.5 \\   

\bottomrule
\end{tabular}
\end{table*}

\begin{table*}[t]
\centering
\small

\begin{minipage}{0.48\linewidth}
\centering
\setlength{\tabcolsep}{3pt}
\caption{BioVITA vs.~TaxaBind.}
\resizebox{\linewidth}{!}{
\begin{tabular}{lcccccccc}
\toprule
& \multicolumn{4}{c}{BioVITA} & \multicolumn{4}{c}{TaxaBench-8k} \\
\cmidrule(lr){2-5}\cmidrule(lr){6-9}
Model
& A$\leftrightarrow$T & A$\leftrightarrow$I & I$\leftrightarrow$T & Avg.
& A$\leftrightarrow$T & A$\leftrightarrow$I & I$\leftrightarrow$T & Avg. \\
\midrule
Taxabind & 30.5 & 38.6 & 83.5 & 50.9 & {\color{gray}22.9} & {\color{gray}30.9} & {\color{gray}47.8} & {\color{gray}33.9} \\
Ours     & {\color{gray}89.3} & {\color{gray}81.5} & {\color{gray}96.8} & {\color{gray}89.2} & 57.0 & 36.4 & 77.2 & 56.9 \\
\bottomrule
\end{tabular}
}
\label{tab:biovita_taxabind_bidirectional}
\end{minipage}
\hfill
\begin{minipage}{0.48\linewidth}
\centering
\setlength{\tabcolsep}{3pt}
\caption{Evaluation on other image and audio benchmarks.}
\resizebox{\linewidth}{!}{
\begin{tabular}{l@{\hspace{3.3pt}}ccccc}
\toprule
Model & CUB-200 & BioCLIP-Rare & iSoundNat \\
\midrule
Taxabind & 75.0 & 34.1 & 16.2 \\
Ours & 91.1 & 82.9 & 44.4 \\
\bottomrule
\end{tabular}
}
\label{tab:baseline_other}
\end{minipage}

\end{table*}


In the main paper, we did not include a comparison with BioLingual \cite{bioacoustics} because its training split may contain samples from our test set.
To avoid this potential data leakage, we construct a new test subset that contains only the 2024 split, which is not included in the BioLingual training data.
This test set consists of 2,710 species and 4,483 recordings.

In this setting, we evaluate classification performance by averaging over classes.
Table~\ref{tab:taxon_species_family_genus} shows the retrieval results using species-, family-, and genus-level prompts, meaning that the input prompt specifies the species name, family name, or genus name of the target class.

The results indicate that, for species-level classification, \scalebox{0.95}{$\mathsf{BioVITA}$} achieves better accuracies compared with BioLingual.
For family- and genus-level prompts, \scalebox{0.95}{$\mathsf{BioVITA}$} clearly outperforms BioLingual.
This suggests that \scalebox{0.95}{$\mathsf{BioVITA}$} benefits from the taxonomy-aware prompting strategy inherited from BioCLIP, enabling the model to generalize more effectively beyond the species level.

\subsection{Evaluation on Other Datasets}
Table~\ref{tab:biovita_taxabind_bidirectional} presents a direct comparison with TaxaBind on species retrieval evaluated by top-5 accuracy. The gray text indicates an in-dataset evaluation. 
Our model consistently outperforms TaxaBind, demonstrating that the larger dataset and VITA training strategy improve retrieval accuracy.

Table~\ref{tab:baseline_other} shows results for top-1 accuracy for zero-shot species-level retrieval averaged over both retrieval directions on CUB-200~\cite{CUB200}, BioCLIP-Rare~\cite{rare_species_2023}, iSoundNat.
Our model demonstrates generality.

\subsection{t-SNE}
\begin{figure*}[t]
\centering
\includegraphics[width=\linewidth]{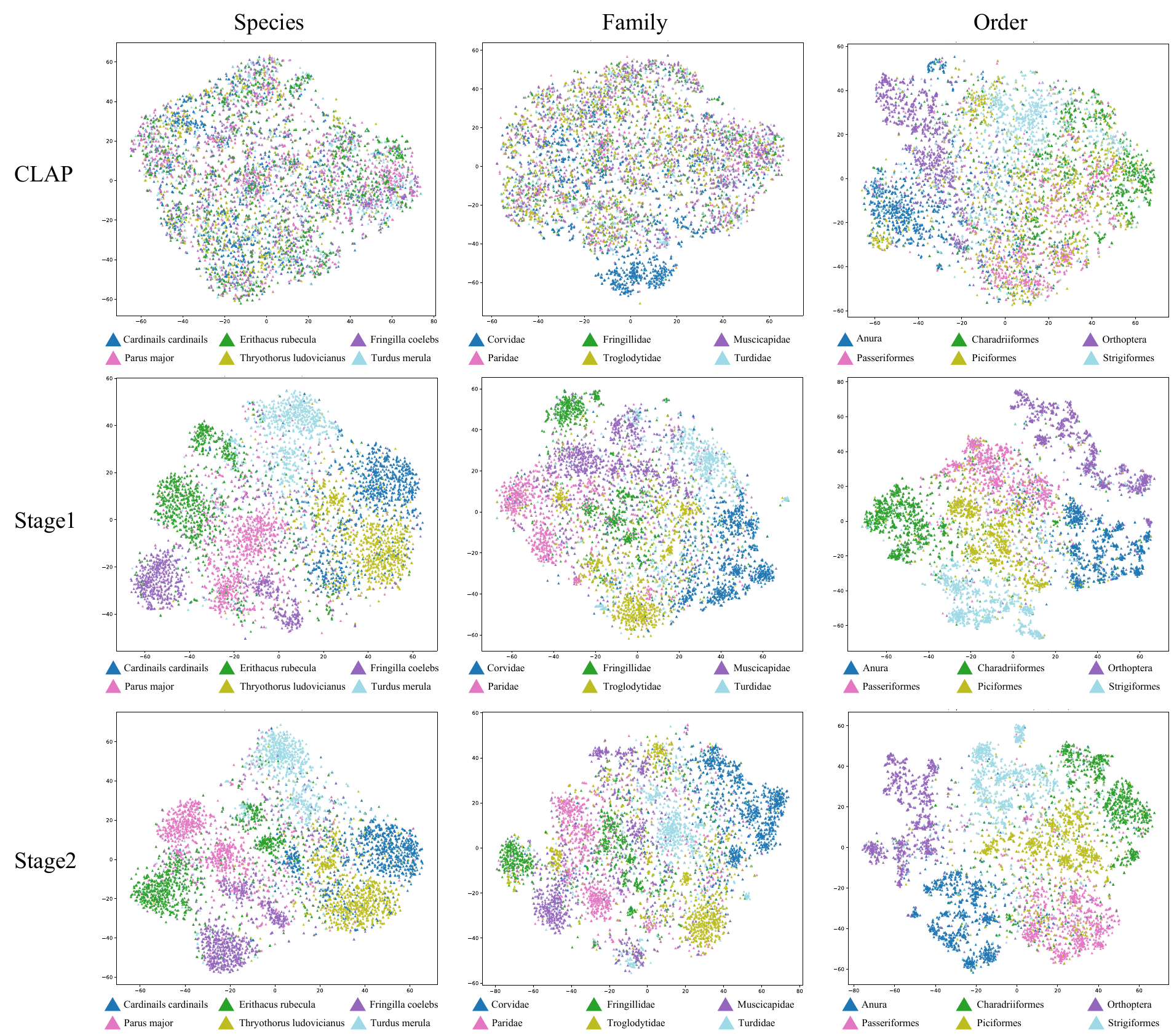}
\vspace{-14pt}
\caption{t-SNE visualization. Our model successfully learns all three modalities in Stage 2 without collapsing the audio feature embedding space.}
\label{fig:tsne}
\vspace{-9pt}
\end{figure*}

We visualize the t-SNE embeddings from the audio encoder in Fig.~\ref{fig:tsne}, clustered by the top six categories in each taxonomy level (species, family, and order). The results show that in Stage 1 (text–audio training), the model learns well-aligned audio–text representations. Moreover, with careful tri-modal learning, Stage 2 successfully preserves the inherent structure of the audio clusters.

\subsection{Training Data Size}
We investigate how the size of the training set affects our model’s performance. In Table~\ref{tab:training_size}, we present results for two reduced dataset settings, one-fourth and one-half of the original training size. If downsampling would remove a species entirely, we retain at least one sample to avoid collapsing the taxonomy structure.
We maintain consistency in the training protocol.

These results demonstrate that the size of the training audio dataset has a substantial impact on model performance. Our \scalebox{0.95}{$\mathsf{BioVITA}$} model benefits greatly from the large-scale dataset and learns robust audio representations from extensive training data.

\begin{table*}[t]
\centering
\small
\setlength{\tabcolsep}{3pt}
\caption{Training dataset size variation. The amount of training data significantly affects model performance. These results correspond to Stage 1. “Sci” and “Com” denote the prompt types used during inference: scientific name and common name, respectively.}
\vspace{-5pt}
\label{tab:training_size}
\begin{tabular}{clcccccccccccccc}
\toprule
\multicolumn{2}{l}{\multirow{2}{*}{\textbf{Model}}}
  & \multicolumn{2}{c}{\textbf{Audio$\rightarrow$Text}} 
  & \multicolumn{2}{c}{\textbf{Text$\rightarrow$Audio}} 
  & \multicolumn{2}{c}{\textbf{Audio$\rightarrow$Image}} 
  & \multicolumn{2}{c}{\textbf{Image$\rightarrow$Audio}} 
  & \multicolumn{2}{c}{\textbf{Image$\rightarrow$Text}} 
  & \multicolumn{2}{c}{\textbf{Text$\rightarrow$Image}}
  & \multicolumn{2}{c}{\textbf{Average}} \\[-0.2em]
\cmidrule(lr){3-4} \cmidrule(lr){5-6} \cmidrule(lr){7-8} \cmidrule(lr){9-10} \cmidrule(lr){11-12} \cmidrule(lr){13-14} \cmidrule(lr){15-16}
\\[-1.3em]
&  & \scalebox{0.95}{\textbf{Top 1}} & \scalebox{0.95}{\textbf{Top 5}}
 & \scalebox{0.95}{\textbf{Top 1}} & \scalebox{0.95}{\textbf{Top 5}} & \scalebox{0.95}{\textbf{Top 1}} & \scalebox{0.95}{\textbf{Top 5}} & \scalebox{0.95}{\textbf{Top 1}} & \scalebox{0.95}{\textbf{Top 5}} & \scalebox{0.95}{\textbf{Top 1}} & \scalebox{0.95}{\textbf{Top 5}} & \scalebox{0.95}{\textbf{Top 1}} & \scalebox{0.95}{\textbf{Top 5}} & \scalebox{0.95}{\textbf{Top 1}} & \scalebox{0.95}{\textbf{Top 5}} \\[-0.2em]
\midrule
\multirow{3}{*}{\rotatebox{90}{\textbf{Sci}}}
& 25\%
& 48.0 & 68.1 & 67.5 & 85.9 & 39.6 & 65.9 & 41.1 & 71.4 & 45.4 & 59.5 & 80.9 & 93.2 & 53.8 & 74.0 \\
& 50\%
& 53.6 & 73.4 & 75.2 & 90.5 & 43.7 & 70.0 & 44.5 & 74.5 & 45.4 & 59.5 & 81.1 & 93.5 & 57.2 & 76.9 \\

& \cellcolor{lightcyan}\scalebox{0.95}{$\mathsf{BioVITA}$} (Full)
 & \cellcolor{lightcyan}60.3 & \cellcolor{lightcyan}80.0 &\cellcolor{lightcyan} 79.3 & \cellcolor{lightcyan}93.9 & \cellcolor{lightcyan}47.8 & \cellcolor{lightcyan}72.8 & \cellcolor{lightcyan}48.6 & \cellcolor{lightcyan}78.8 &\cellcolor{lightcyan} 65.1 & \cellcolor{lightcyan}80.5 &\cellcolor{lightcyan} 84.8 & \cellcolor{lightcyan}95.3 &\cellcolor{lightcyan} 64.3 & \cellcolor{lightcyan}83.6  \\
[-0.2em]
\midrule
\multirow{3}{*}{\rotatebox{90}{\textbf{Com}}}
& 25\%
& 50.0 & 72.1 & 65.8 & 86.7 & 39.6 & 65.9 & 41.1 & 71.4 & 65.1 & 80.4 & 84.5 & 95.1 & 57.7 & 78.6 \\
& 50\%
 & 55.7 & 76.3 & 73.0 & 90.6 & 43.7 & 70.0 & 44.5 & 74.5 & 65.1 & 80.5 & 84.6 & 95.2 & 61.1 & 81.2 \\

& \cellcolor{lightcyan}\scalebox{0.95}{$\mathsf{BioVITA}$} (Full)
& \cellcolor{lightcyan}60.4 &\cellcolor{lightcyan} 80.0 &\cellcolor{lightcyan} 79.3 & \cellcolor{lightcyan}93.9 & \cellcolor{lightcyan}48.0 & \cellcolor{lightcyan}72.8 & \cellcolor{lightcyan}48.8 & \cellcolor{lightcyan}78.9 &\cellcolor{lightcyan} 65.2 & \cellcolor{lightcyan}80.5 & \cellcolor{lightcyan}84.7 & \cellcolor{lightcyan}95.0 & \cellcolor{lightcyan}64.4 & \cellcolor{lightcyan}83.5 \\
[-0.2em]

\bottomrule
\end{tabular}
\end{table*}

\section{Dataset Details}
\subsection{Annotation Example}
\begin{figure*}
\centering
\includegraphics[width=0.98\linewidth]{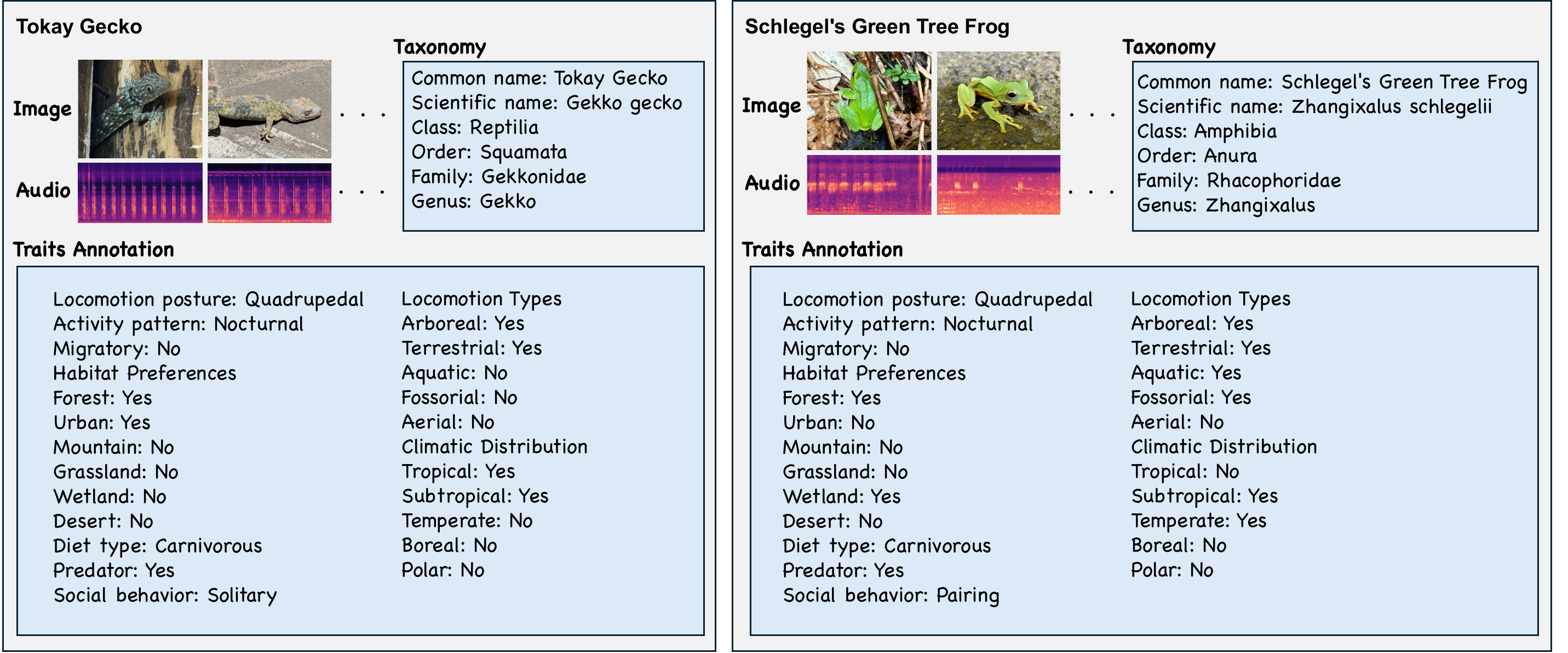}
\vspace{-5pt}
\caption{Dataset Example.}
\label{fig:dataset_ex_1}

\end{figure*}

We present annotation examples for Tokay Gecko 
and Schlegel's Green Tree Frog
in Fig.~\ref{fig:dataset_ex_1}. For each animal species, we collect images, audio recordings, taxonomic information, and trait annotations.

\subsection{Dataset Distribution}
\begin{figure*}
\centering
\includegraphics[width=0.81\linewidth]{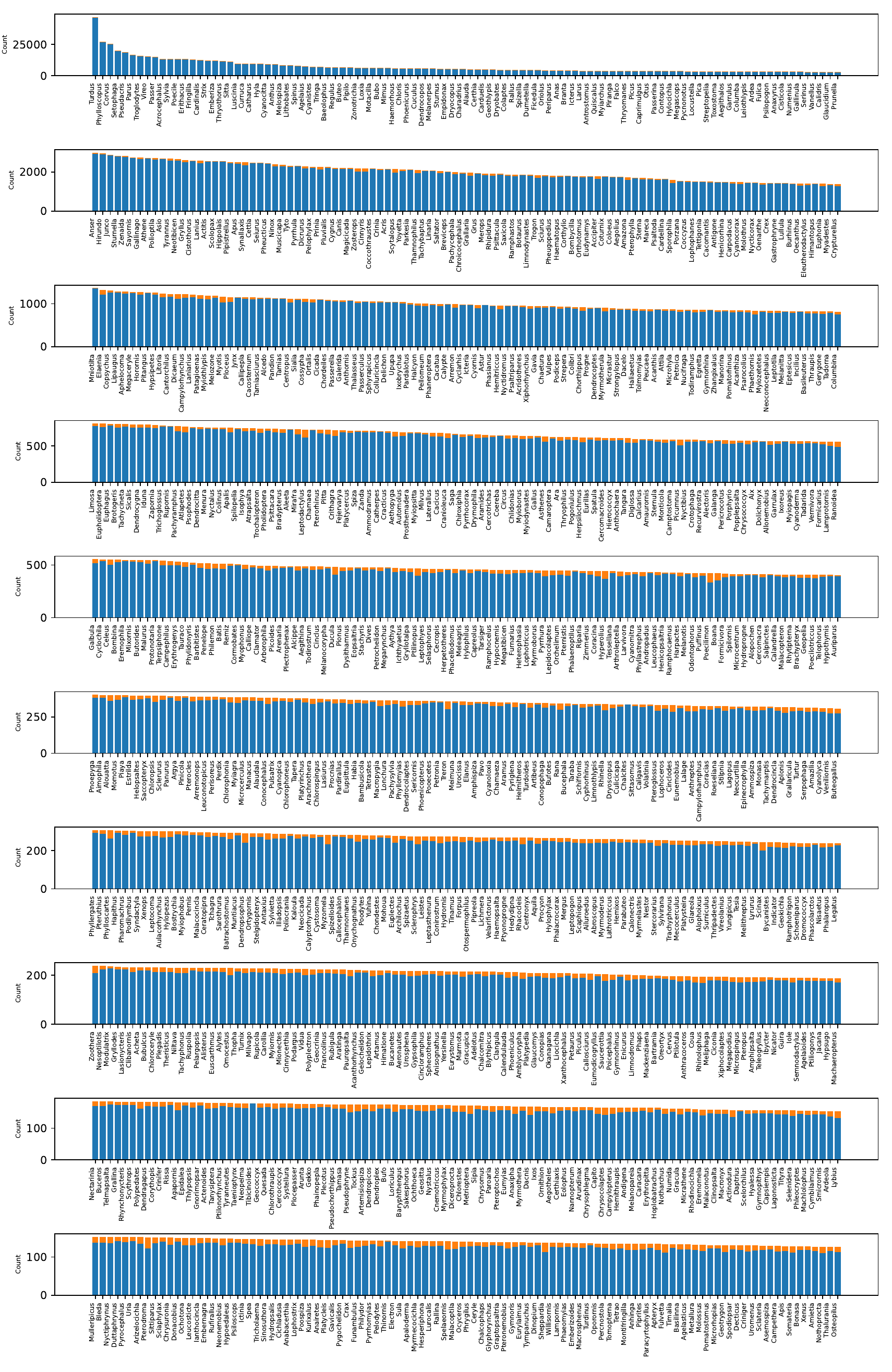}
\caption{Genus distribution. The blue bars represent the training set, while the orange bars represent the test set.}
\label{fig:genus1}
\end{figure*}

\begin{figure*}
\centering
\includegraphics[width=0.79\linewidth]{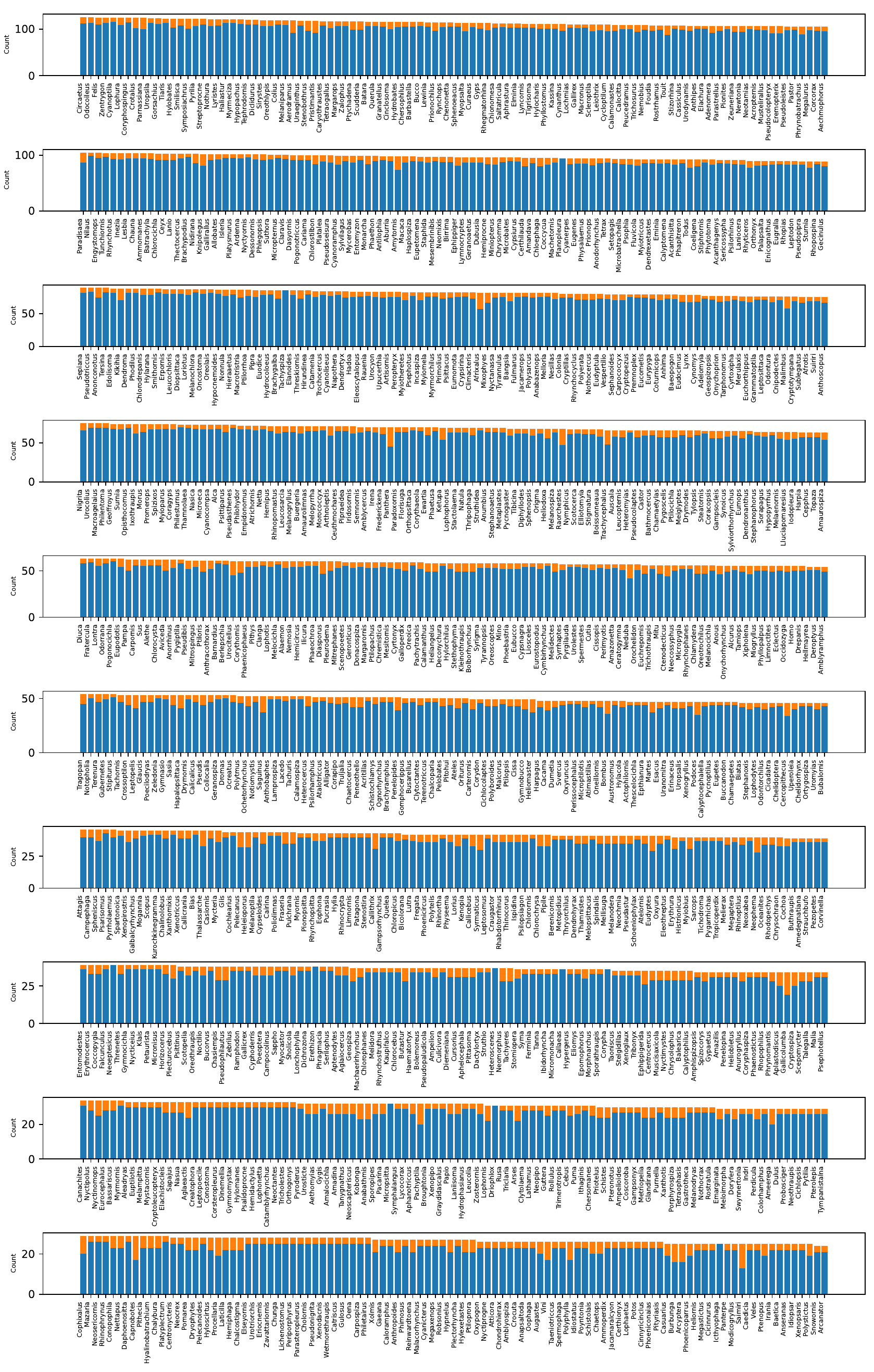}
\caption{Genus distribution. The blue bars represent the training set, while the orange bars represent the test set.}
\label{fig:genus2}
\end{figure*}

\begin{figure*}
\centering
\includegraphics[width=0.81\linewidth]{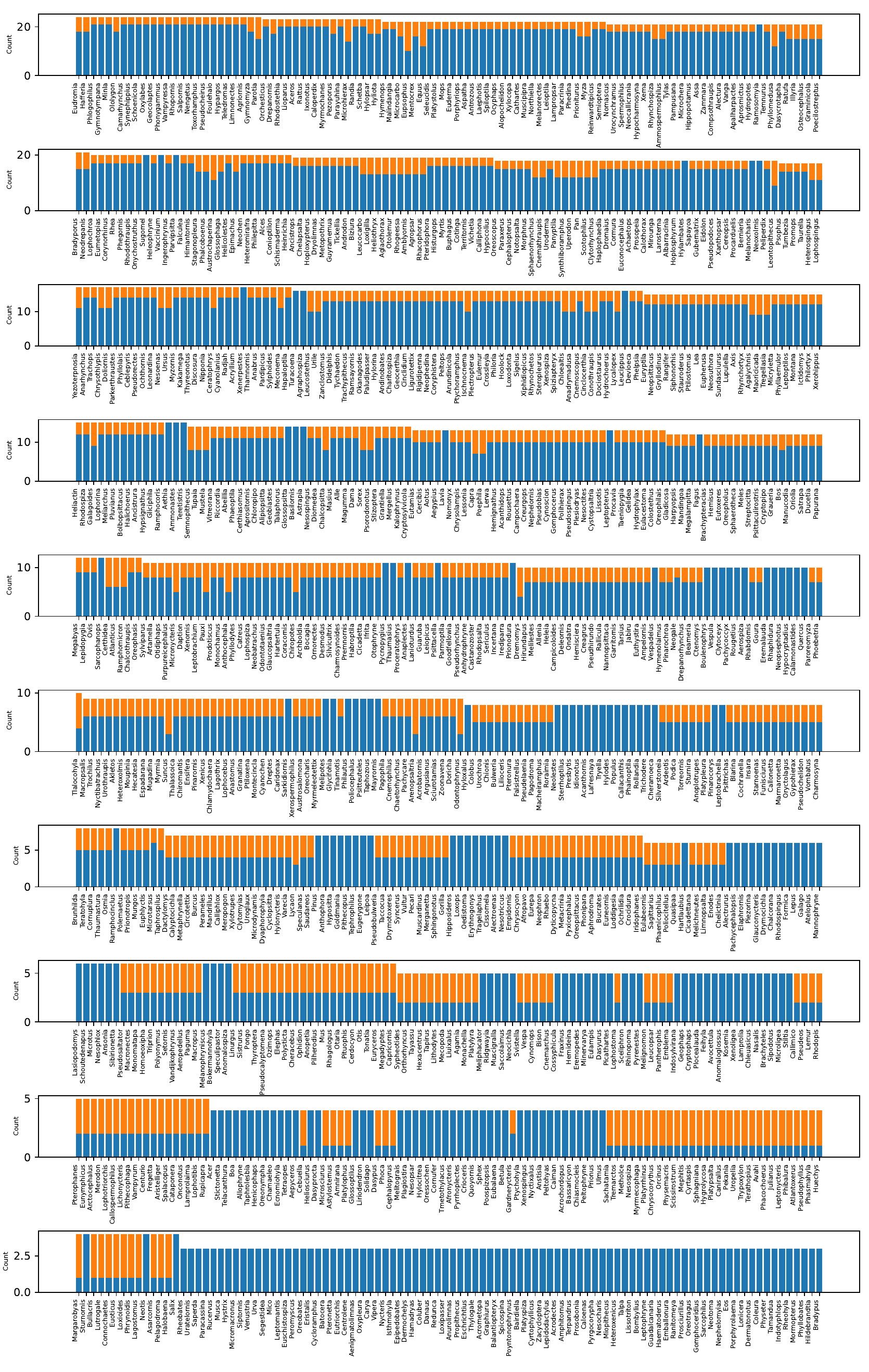}
\caption{Genus distribution. The blue bars represent the training set, while the orange bars represent the test set.}
\label{fig:genus3}
\end{figure*}

\begin{figure*}
\centering
\includegraphics[width=0.85\linewidth]{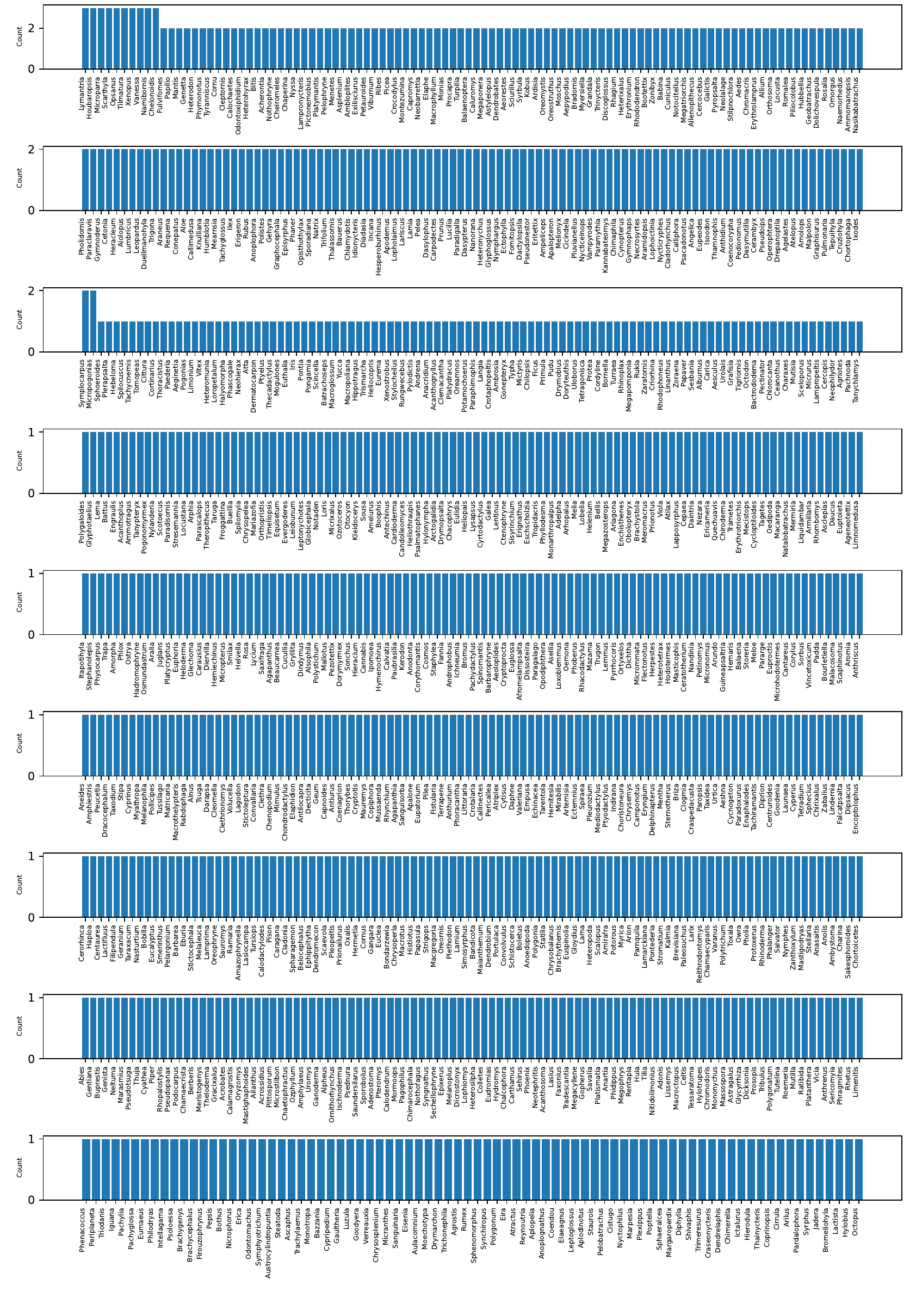}
\caption{Genus distribution. The blue bars represent the training set, while the orange bars represent the test set.}
\label{fig:genus4}
\end{figure*}

We illustrate the genus-level distribution of our dataset in Fig.~\ref{fig:genus1} through Fig.~\ref{fig:genus4}.
The blue bars represent the training set, while the orange bars represent the test set.

\clearpage
\twocolumn

\end{document}